\providecommand{\GCEACMOptions}{manuscript,nonacm}
\documentclass[\GCEACMOptions]{acmart}

\usepackage{amsmath,amsfonts}
\usepackage{array}
\usepackage{booktabs}
\usepackage{multirow}
\usepackage{graphicx}
\usepackage{bm}
\usepackage{mathrsfs}
\usepackage{rotating}
\usepackage{url}

\usepackage{amsmath,amsfonts,bm}

\def\eqref#1{equation~\ref{#1}}

\def\1{\bm{1}}

\DeclareMathAlphabet{\mathsfit}{\encodingdefault}{\sfdefault}{m}{sl}
\SetMathAlphabet{\mathsfit}{bold}{\encodingdefault}{\sfdefault}{bx}{n}

\setcopyright{none}
\newcommand{\LL}{\mathcal{L}}
\newcommand{\EE}{\mathop{\mathbb{E}}}
\newlength{\imsize}
\newlength{\sgsize}
\newlength{\textsize}
\newlength{\xsize}
\newlength{\figup}
\newlength{\figmid}
\newlength{\figdwn}
\newlength{\tabup}
\newlength{\tabmid}
\newlength{\tabdwn}
\title[Generated Contents Enrichment]{Generated Contents Enrichment}

\author{Mahdi Naseri}
\affiliation{%
  \institution{University of Waterloo}
  \department{Department of Electrical and Computer Engineering}
  \city{Waterloo}
  \state{ON}
  \country{Canada}}
\email{mahdi.naseri@uwaterloo.ca}

\author{Jiayan Qiu}
\affiliation{%
  \institution{University of Waterloo}
  \department{Department of Electrical and Computer Engineering}
  \city{Waterloo}
  \state{ON}
  \country{Canada}}
\email{jiayan.qiu.1991@outlook.com}

\author{Zhou Wang}
\affiliation{%
  \institution{University of Waterloo}
  \department{Department of Electrical and Computer Engineering}
  \city{Waterloo}
  \state{ON}
  \country{Canada}}
\email{zhou.wang@uwaterloo.ca}

\begin{abstract}
    We study Generated Contents Enrichment (GCE), a conditional image-generation task in which a sparse scene description is first enriched through an explicit scene representation and then rendered into semantically richer visual content. Conventional image-generation systems can produce visually realistic outputs from limited scene descriptions, but the added content is usually implicit in the generator rather than represented as an inspectable intermediate structure. In contrast, GCE seeks to make scene enrichment explicit at the scene-representation level while examining its visual consequences during generation, with the goal of encouraging generated content that is visually plausible, structurally coherent, and semantically richer than the sparse input.
    To instantiate GCE, we propose a jointly trained adversarial framework that enriches scene graphs by modeling object semantics and inter-object relations. Our approach first represents the input description as a scene graph, where nodes model objects and edges capture inter-object relations. The framework uses graph convolutional networks to predict additional objects and their relations to the existing scene. Finally, the enriched scene graph is passed through the downstream image-generation pipeline to generate the corresponding visual content. We evaluate the framework with proxy scene graph enrichment metrics, image-quality comparisons, qualitative examples, and user studies on the Visual Genome dataset.
\end{abstract}



\hypersetup{
  pdfauthor={Mahdi Naseri, Jiayan Qiu, and Zhou Wang},
  pdftitle={Generated Contents Enrichment},
  pdfsubject={Public preprint},
  pdfkeywords={Generated Contents Enrichment, scene graph enrichment, graph-conditioned image generation, image synthesis, graph convolutional networks}
}

\newcommand{\GCEPlainPageStyles}{%
  \fancypagestyle{firstpagestyle}{%
    \fancyhf{}%
    \fancyfoot[C]{\footnotesize\thepage}%
    \renewcommand{\headrulewidth}{0pt}%
    \renewcommand{\footrulewidth}{0pt}%
  }%
  \fancypagestyle{standardpagestyle}{%
    \fancyhf{}%
    \fancyfoot[C]{\footnotesize\thepage}%
    \renewcommand{\headrulewidth}{0pt}%
    \renewcommand{\footrulewidth}{0pt}%
  }%
}

\begin{document}
\GCEPlainPageStyles
\maketitle
\GCEPlainPageStyles
\thispagestyle{firstpagestyle}
\pagestyle{standardpagestyle}
\section{Introduction}
\begin{figure*}
    \vspace{\figup}
    \centering
    \raisebox{0.5\xsize}{
        \begin{minipage}[c]{0.65\xsize}
            \centering \scriptsize
            The floor is under the sink. The floor has tile. The wall has a frame.
        \end{minipage}}
    \raisebox{0.5\xsize}{
        \begin{minipage}[c]{\xsize}
            \centering
            \includegraphics[width=\xsize]{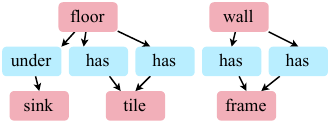}
        \end{minipage}}
    \raisebox{0.5\xsize}{
        \begin{minipage}[c]{1.35\xsize}
            \centering
            \includegraphics[width=1.35\xsize]{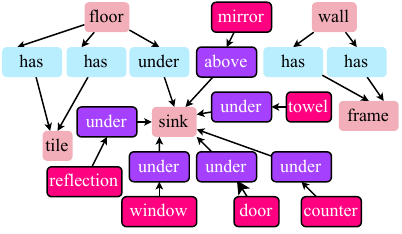}
        \end{minipage}}
    \includegraphics[width=\xsize]{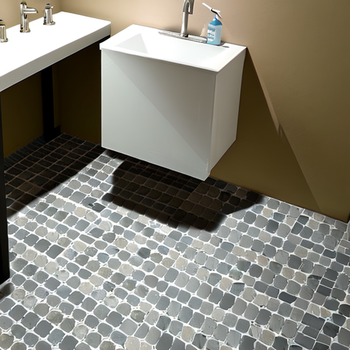}\hspace{-0.5mm}
    \includegraphics[width=\xsize]{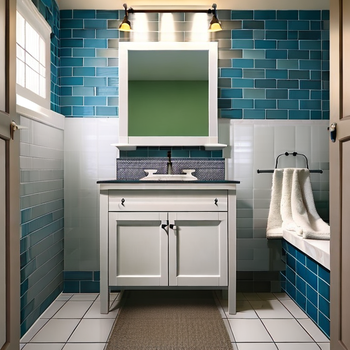}\hspace{-0.5mm}
    \includegraphics[width=\xsize]{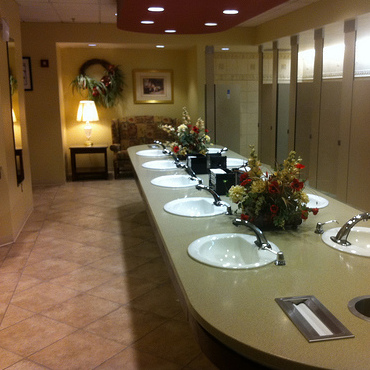}\\
    \vspace{0.25em}
    \raisebox{0.5\xsize}{
        \begin{minipage}[c]{0.65\xsize}
            \centering \scriptsize
            A tree and a street. A bush is next to the sidewalk.
        \end{minipage}}
    \raisebox{0.5\xsize}{
        \begin{minipage}[c]{\xsize}
            \centering
            \includegraphics[width=\xsize]{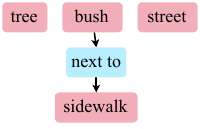}
        \end{minipage}}
    \raisebox{0.5\xsize}{
        \begin{minipage}[c]{1.35\xsize}
            \centering
            \includegraphics[width=1.35\xsize]{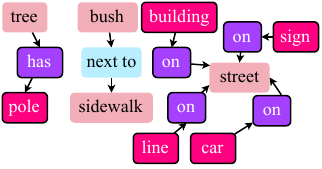}
        \end{minipage}}
    \includegraphics[width=\xsize]{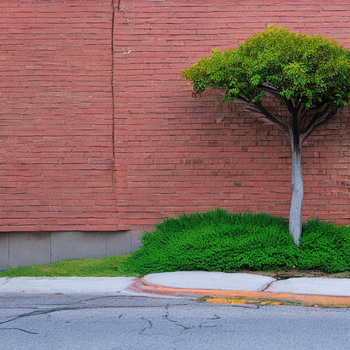}\hspace{-0.5mm}
    \includegraphics[width=\xsize]{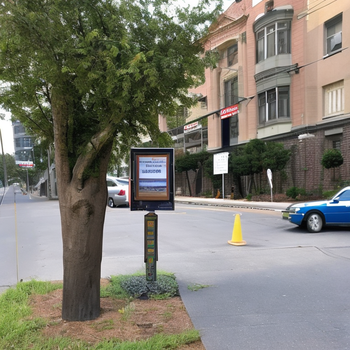}\hspace{-0.5mm}
    \includegraphics[width=\xsize]{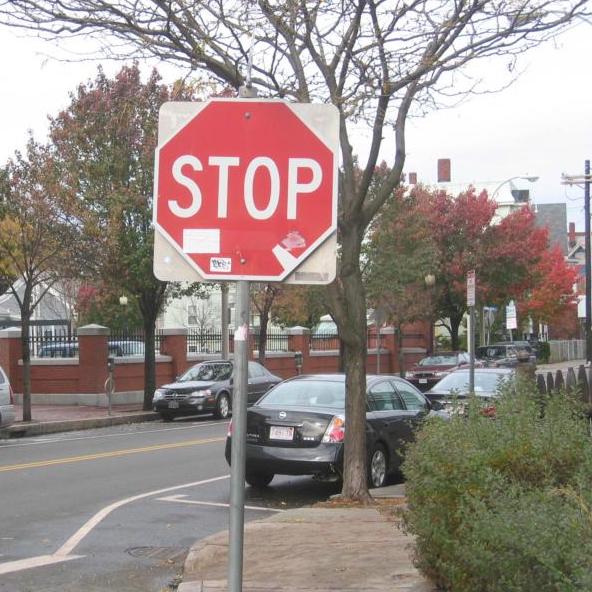}\\
    \vspace{-0.5mm}
    \begin{minipage}{0.8\xsize} \centering\scriptsize (a) Input Description \end{minipage}
    \begin{minipage}{\xsize} \centering\scriptsize (b) Input Scene Graph \end{minipage}
    \begin{minipage}{1.3\xsize} \centering\scriptsize (c) Enriched Scene Graph \end{minipage}
    \begin{minipage}{\xsize} \centering\scriptsize (d) Simple Generated \end{minipage}
    \begin{minipage}{\xsize} \centering\scriptsize (e) Enriched Generated \end{minipage}
    \begin{minipage}{\xsize} \centering\scriptsize (f) Real-World \end{minipage}\\
    \vspace{\figmid}
    \caption{\emph{Overview of GCE.} The sparse input description (a) is converted to a scene graph (b). The graph is enriched with additional objects and relations (c), then rendered as an enriched image (e). The added object and relation semantics should preserve the essential scene characteristics of the input graph. Our adversarial graph-convolutional enrichment framework performs this structured enrichment before image synthesis. The input graph (b) and enriched graph (c) produce the simple image (d) and enriched image (e), respectively. Compared with (d), the enriched image (e) preserves the input scene while adding object and relation semantics closer to the real Visual Genome image (f).}
    \label{fig:teaser}
    \vspace{0.5em}
\end{figure*}

Real-world scenes, such as those in Fig.~\ref{fig:teaser}(f), contain rich objects and relations. The short descriptions in Fig.~\ref{fig:teaser}(a) are often inadequately informative, creating a semantic richness gap between the simple generated images in Fig.~\ref{fig:teaser}(d) and the real scenes in Fig.~\ref{fig:teaser}(f). However, given a sparse description, humans can often imagine a semantically richer scene by understanding the description, adding related semantics and inter-semantic relations, and visualizing the resulting content.

We define Generated Contents Enrichment (GCE) as a scene-representation enrichment task for image generation: given a sparse text description, the system constructs a scene graph, enriches it with plausible additional objects and relations, and renders visual content from the enriched representation.
Neighboring graph-expansion methods, such as GEMS \citep{agarwal2023gems}, expand scene structures for retrieval- or language-domain objectives, while graph-to-image methods, such as SG2IM \citep{johnson2018image}, synthesize images from already specified scene graphs. In contrast, GCE studies explicit scene enrichment as a structured intermediate stage for image generation; our framework is one instantiation of this task that evaluates the visual consequences of added objects and relations during training.
Compared with scene graph completion, unconditional scene graph generation, graph-to-image rendering, or large language model (LLM) prompt expansion, GCE is defined by the full path from a sparse description and its seed graph to an enriched graph and then to rendered visual content.

The method is not only asked to fill a missing graph element or render an already specified graph. This requirement is part of the GCE definition: the enriched graph should add object-relation semantics that are plausible for the scene, compatible with the seed description, and useful for downstream image synthesis. This explicit graph-level enrichment step makes the model's added scene content inspectable before rendering, while the human analogy serves only as motivation for the task.
Table~\ref{tab:gce_neighboring_tasks} summarizes these distinctions; the key point is that GCE exposes graph-level enrichment before image synthesis, whereas neighboring settings either complete graphs, generate graphs, render specified graphs, or expand text prompts. Scene graph completion predicts missing graph elements inside a graph-only problem. Scene graph generation produces a graph, often without conditioning on the same sparse seed description and downstream image objective. Graph-to-image rendering assumes the conditioning graph is already specified. Prompt expansion enriches text but does not expose an explicit graph-level enrichment output. GCE combines these pressures: it starts from a sparse scene description represented as a reduced scene graph, predicts additional object-relation semantics as an inspectable intermediate representation, and evaluates the result through both graph-side and image-side evidence.

\begin{table*}[t]
    \centering
    \vspace{0.2em}
    \caption{\emph{GCE compared with neighboring settings.} GCE is distinguished by explicit graph-level enrichment before rendering and by evaluation of both the enriched graph and the rendered image.}
    \label{tab:gce_neighboring_tasks}
    \vspace{\tabmid}
    \small
    \resizebox{\textwidth}{!}{%
        \begin{tabular}{llll}
            \toprule
            \textbf{Setting}         & \textbf{Input}                                          & \textbf{Output}                   & \textbf{Role relative to this paper}                                  \\ \midrule
            Scene graph completion   & Partial graph                                           & Completed graph                   & Nearby graph-only recovery task.                                      \\
            Scene graph generation   & Dataset prior or conditioning signal                    & Generated graph                   & Nearby graph-generation setting without the same rendering objective. \\
            Graph-to-image rendering & Specified graph                                         & Image                             & Downstream rendering task once graph content is already given.        \\
            LLM prompt expansion     & Text prompt                                             & Expanded text prompt              & Text-space enrichment without explicit object-relation graph output.  \\
            GCE                      & Sparse description represented as a reduced scene graph & Enriched graph and rendered image & Target setting studied in this paper.                                 \\ \bottomrule
        \end{tabular}%
    }
    \vspace{\tabdwn}
\end{table*}

Although humans can readily infer richer scenes from brief descriptions, existing image-generation methods \citep{karras2020analyzing, vahdat2020nvae, chen2020generative, saharia2022photorealistic, rombach2022high, esser2021taming, yang2022diffusion, herzig2020learning} may add unstated content, but they usually do not expose or evaluate this enrichment as an explicit structured step. In contrast, our framework explicitly enriches the scene representation by reasoning about additional relevant semantics and inter-semantic relations. As illustrated in Fig.~\ref{fig:teaser}(d)--(f), standard generation pipelines can produce visually plausible images, yet their outputs often remain semantically less rich than the corresponding real-world scenes. The enriched generation in Fig.~\ref{fig:teaser}(e) instead makes the added scene content explicit before rendering.

The semantic richness gap in Fig.~\ref{fig:teaser} motivates GCE, but learning this enrichment is challenging. First, the enriching content should be semantically relevant to the described content and should form appropriate inter-semantic relations. Second, the enriched scene graph should remain structurally plausible under the distribution of scene graphs observed in real images.

To address these challenges, we propose an approach inspired by the way humans reason about and imagine plausible enriching content. Our framework makes this human-inspired enrichment step explicit in a scene graph before image generation. Our method has a jointly trained architecture with three principal stages. Before the three learned stages, the input description is represented as a scene graph, where each node denotes an object and each edge denotes an inter-object relation. In the first stage, we enrich the scene graph iteratively. In each iteration, we add one temporary object slot and predict both its object label and its incident relation labels. These predictions are produced by a Graph Convolutional Network (GCN), which aggregates information from known nodes and edges. As a result, the enriching content remains semantically aligned with the content in the input description. The enriched scene graph is then scored by scene graph critics that encourage realistic object and relation structure. In the second stage, the enriched graph is converted into a continuous text-conditioning representation and rendered by a fixed text-to-image generator. During training, object and predicate predictions are kept as soft probability distributions and mapped through the corresponding category or token embeddings. This continuous conditioning path allows visual and image-text alignment losses to backpropagate through the frozen conditioning pathway while updating only the scene graph enrichment modules. In the third stage, a visual scene characterizer and an image-text aligner provide training signals that encourage the generated image to remain plausible and consistent with the input description.

Our main contribution is a jointly trained framework that instantiates GCE through explicit scene graph enrichment before image generation. The GCE task aims to narrow the semantic richness gap between sparse scene descriptions and the richer scenes that humans can readily infer from them. Conventional text-to-image generation approaches mainly rely on implicit enrichment within the generator. In contrast, our framework explicitly enriches the scene representation while using visual-domain signals to guide the generation of richer output images. It predicts new objects and relations, evaluates their visual effects during training, and encourages semantic and structural coherence with the input description. The final enriched image preserves the key input elements while adding semantically and structurally coherent content. We evaluate the framework on Visual Genome \citep{krishna2017visual} using proxy scene graph enrichment metrics, image-quality comparisons, qualitative examples, and user studies.

\section{Related Work}

\subsection{Realistic Image Generation}
Generative Adversarial Networks (GANs) \citep{goodfellow2020generative, radford2015unsupervised} jointly train a generator and discriminator to distinguish real from fake synthesized images. GANs can produce high-quality images that are sometimes difficult for humans to distinguish from real images \citep{brock2018large, karras2020analyzing, mirza2014conditional, odena2017conditional}. However, their training procedure can be less stable than alternative generative methods, and they may not model all parts of the data distribution \citep{metz2016unrolled}. Early GAN-based text-to-image synthesis methods \citep{reed2016generative, reed2016learning} were followed by stacked GANs \citep{zhang2018stackgan++} and attention-based models \citep{xu2018attngan, bahdanau2014neural, vaswani2017attention}.

Variational autoencoders (VAEs) \citep{kingma2013auto} map images to latent spaces and decode them back to the image domain. They have also been used for image synthesis \citep{vahdat2020nvae, huang2018introvae}, generating high-resolution data but often with lower perceptual quality than GAN counterparts.

Alternatively, autoregressive approaches \citep{van2016pixel, van2016conditional, chen2020generative, child2019generating, yu2022scaling} synthesize pixels sequentially, conditioned on an encoded input description and previously generated pixels, which can make training and sampling computationally expensive. Although recent diffusion-based models \citep{sohl2015deep, dhariwal2021diffusion, ho2020denoising, song2020score, saharia2022photorealistic} synthesize high-quality images, they often require substantial training cost and iterative inference. Stable Diffusion \citep{rombach2022high} popularized open-source latent diffusion by applying the diffusion process in a lower-dimensional latent image space. More recent systems further scale rectified-flow transformers and unified or instruction-conditioned image generation \citep{esser2024scaling, xiao2025omnigen, chen2025januspro}. Relation-aware prompting also improves text-to-image generation by targeting object-relation control \citep{ye2025rat2igen}. Hybrid two-stage methods such as VQ-VAEs \citep{razavi2019generating, yan2021videogpt} and VQGANs \citep{esser2021taming, yu2021vector} instead combine discrete latent representations with generative modeling.
\subsection{Scene Graphs}

Scene graphs proposed in \cite{al.2015image} and further investigated in \cite{tripathi2021sg2caps, al.2021kimera, zhang2019graphical} are structured scene representations that contain objects, relations, and often attributes. Previously, understanding a scene was limited to detecting and recognizing the objects in an image. However, scene graphs aim to obtain a higher level of scene understanding \citep{chang2021comprehensive} and are employed to improve
action recognition \citep{aksoy2010categorizing},
image captioning \citep{aditya2015from},
and image retrieval \citep{johnson2015image, liu2024semscene}.
Scene graphs may be predicted from images \citep{lu2016visual, xu2017scene, yang2018graph, tang2020unbiased, yang2022panoptic, li2018factorizable, sun2023boosting, li2023zeroshot, li2024decomposed, xu2025sceneadaptive}
or serve as inputs to generate synthesized images \citep{johnson2018image, yang2022diffusion, herzig2020learning, li2019pastegan, ashual2019specifying, wang2024scenegraphdc, wang2025scenegraphgrounded, vo2026simgraph}.
These scene graph to image methods condition on a provided graph, whereas our framework instantiates GCE by inserting an explicit enrichment stage before image synthesis and then studies the visual effect of the added objects and relations.

Generic graph-generation methods provide adjacent architectural context, including molecular graph generators that model graph distributions under chemical validity and property constraints \citep{simonovsky2018graphvae, you2018graphrnn, shi2020graphaf, luo2021graphdf, de2018molgan, goyal2020graphgen, grover2019graphite, jin2018junction, samanta2020nevae}. These methods are useful graph-generation context, but they are not direct task-level counterparts to GCE: molecular graphs are typically smaller, use different validity constraints, and often assume graph properties such as connectivity that do not necessarily hold for scene graphs \citep{agarwal2023gems}. By contrast, GCE enriches scene graphs under scene-semantic and image-consistency constraints before rendering.

\subsection{Scene Graph Enrichment and Graph-Conditioned Generation}
We position GCE as a scene-enrichment task for image generation and discuss our method as one instantiation of that task.
Image-conditioned scene graph enrichment works \citep{khan2023neusyre, khan2022expressive} start from an input image and produce a richer scene graph for visual understanding. They enrich the graph using cues tied to the observed image and its inferred context. Likewise, the outpainting method in \cite{yang2022scene} also starts from an image and extends its scene graph to support spatial extrapolation. This differs from GCE, where enrichment is introduced before image synthesis rather than inferred from an observed image. These image-based methods therefore do not address the same setting as GCE, where enrichment is introduced before image synthesis rather than derived from an observed image.

GEMS \citep{agarwal2023gems} studies scene expansion from a seed graph, primarily evaluating graph-side expansion and retrieval-style plausibility rather than downstream image generation. Our framework evaluates the added objects and relations both as graph outputs and through their effect on the downstream image. Taken together, these neighboring directions differ from GCE in objective, input-output structure, and evaluation emphasis.

SceneGraphGen \citep{scenegraphgen2021} is an unconditional scene graph generation method that can complete a partial graph through its generation interface. It is not designed as a GCE system, where the input is a sparse seed graph and the evaluation asks for compatible enriched structure under a reduced-graph protocol. We use an adapted SceneGraphGen+ baseline in the experiments because it is the most practical graph-generation method we found with an interface that can be aligned to the reduced-graph GCE protocol.

Image-conditioned scene graph generation methods infer graph structure from observed images \citep{zellers2018neural, lin2020gps-net, yang2022panoptic, khan2023neusyre}. Scene expansion and graph-driven data-synthesis methods extend or sample structured scene content \citep{yang2022scene, agarwal2023gems, gao2026generateanyscene, wang2026syntheticcurriculum}, but they do not evaluate the full path from sparse description to enriched representation to rendered image. Graph-to-image methods instead assume that the conditioning graph is already specified \citep{yang2022diffusion, johnson2018image, wang2025scenegraphgrounded, wang2024scenegraphdc, vo2026simgraph}. GCE differs from these neighboring settings by treating graph-level enrichment as an explicit intermediate step for image generation, so the added content can be evaluated before and after rendering.

\section{Method}

\begin{figure*}
    \vspace{\figup}
    \centering
    \includegraphics[width=\textwidth]{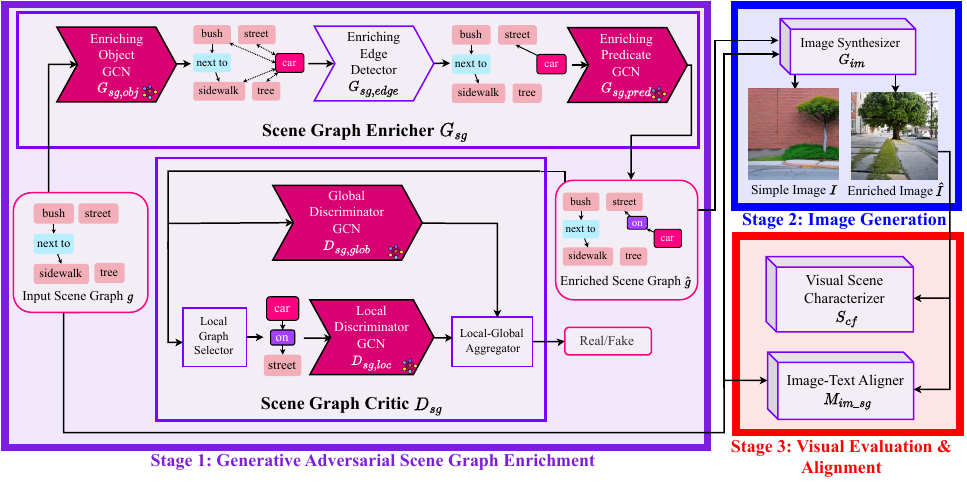}
    \vspace{\figmid}
    \vspace{-6mm}
    \caption{
        \emph{High-Level Overview} of the proposed GCE framework during training. In Stage 1, the input scene graph is fed to the \emph{Scene Graph Enricher $G_{sg}$}, which produces an enriched scene graph. The \emph{Scene Graph Critic $D_{sg}$}, implemented as a pair of local and global discriminators, provides adversarial training signals on the enriched scene graph. These training signals encourage the enriched scene graph to remain structurally coherent and semantically plausible. In Stage 2, the \emph{Image Synthesizer $G_{im}$} renders an image from the enriched scene graph. In Stage 3, the \emph{Visual Scene Characterizer $S_{cf}$} and the \emph{Image-Text Aligner $M_{im\_sg}$} extract visual and text-alignment signals from the rendered image and the original scene information. Stage 3 is used only during training to encourage consistency; at inference, the learned enricher and renderer produce the output.
    }
    \label{fig:pipeline}
    \vspace{\figdwn}
\end{figure*}

We first define GCE at the task level, then instantiate it with the graph-based framework in Fig.~\ref{fig:pipeline}. At the task level, GCE maps a sparse scene description $x$ to an enriched scene representation and a rendered image, making semantic enrichment explicit before image synthesis. Our framework instantiates this task as \(g=T(x)\), \(\hat g=G_{sg}(g)\), and \(\hat I=G_{im}(\hat g)\), where \(T\) constructs the reduced input graph \(g\), \(G_{sg}\) enriches it, and \(G_{im}\) renders the enriched graph. Here, $g$ denotes the constructed scene graph and $\hat g$ its enriched version, so the framework makes the task-level enrichment step concrete as $g \rightarrow \hat g$.

This framework operates as a three-stage pipeline: Stage 1 enriches $g$ into $\hat g$, Stage 2 renders images from scene graphs, and Stage 3 provides training-time visual alignment signals.
First, \emph{Stage 1: Scene Graph Enrichment} performs content enrichment at the object and relation level using GCNs. The \emph{Scene Graph Enricher} uses GCNs and multilayer perceptrons (MLPs) to predict an added object and its relations to the existing scene graph. The \emph{Scene Graph Critic}, implemented as a pair of local and global discriminators, encourages the enriched parts to remain realistic, structurally coherent, and semantically meaningful both on their own and within the context of the full scene graph. The enriched scene graph then proceeds to \emph{Stage 2: Image Generation}, where it serves as the basis for synthesizing an enriched image containing richer relevant content. Lastly, the enriched image, together with the original scene information, is passed to \emph{Stage 3: Training-Time Visual Alignment}. This stage does not define the task output; instead, it provides training-time signals that encourage the generated image to remain visually plausible and consistent with the original scene characteristics. These signals are provided by the scene classifier and the image/text encoders introduced below.
At inference time, the framework uses the learned scene graph enricher to produce \(\hat g\), then renders the image from \(\hat g\). The \emph{Scene Graph Critic $D_{sg}$} and the Stage 3 modules are training-only components and are detached once GT-based supervision is unavailable.
\subsection{Scene Graphs}

The input scene graph is \(g=(O,E)\). The object set is \(O=\{o_1,\ldots,o_n\}\), and the directed edge set is \(E\subseteq O\times\mathcal{R}\times O\), where \(\mathcal{R}\) denotes the relation-category vocabulary. Each object is represented by a category \(o_i\in\mathcal{C}\), where \(\mathcal{C}\) is the set of object categories. Each edge has the form \((o_i,r,o_j)\), where \(o_i,o_j\in O\) and \(r\in\mathcal{R}\) describes the relation type. In this directed triplet, \(o_i\) is the subject, \(o_j\) is the object, and \(r\) is the predicate.
For an iterative enrichment step, let \(g_t=(O_t,E_t)\) denote the current graph using the same object and edge notation. The scene graph enricher predicts an added object \(\hat o_t\in\mathcal{C}\) and incident relations \(\hat E_t\subseteq (O_t\times\mathcal{R}\times\{\hat o_t\})\cup(\{\hat o_t\}\times\mathcal{R}\times O_t)\). The next graph is
\begin{equation}
    g_{t+1}=(O_t\cup\{\hat o_t\},\,E_t\cup\hat E_t).
    \label{eq:gce_iterative_update}
\end{equation}

\subsection{Graph Convolutional Network (GCN)}

Our GCE architecture uses SG2IM-style graph convolution \citep{johnson2018image} as a building block, but adapts it to predict new scene content rather than only render a fully specified graph. We use \emph{graph convolutional network} (GCN) for the full stacked network and \emph{graph convolution} (GConv) for one message-passing layer inside that stack. Following the SG2IM-style design with our local modifications, each GConv jointly updates subject, predicate, and object representations from relation triplets. In this framework, the $l^{th}$ GConv layer, denoted as $GConv^{(l)}$ performs the message-passing as:
\begin{equation}
    \begin{split}
        (V_O^{(l+1)}, V_R^{(l+1)}) = GConv^{(l)}(V_O^{(l)}, V_R^{(l)}, O, E).
    \end{split}
\end{equation}
In the context of the $GConv^{(l)}$'s input graph, $V_O^{(l)}$ represents the feature vectors for all nodes, and $V_R^{(l)}$ denotes the feature vectors for all predicates. Additionally, the corresponding output vectors are characterized as $V_O^{(l + 1)}$ and $V_R^{(l + 1)}$. To initialize the process, $V_O^{(0)}$ and $V_R^{(0)}$ serve as the initial vector values for the objects and predicates within the GCN input graph.

The fundamental building block of our GCN is the GConv layer. For a triplet \((o_i,r,o_j)\) at layer \(l\), the vectors \(v_i\), \(v_r\), and \(v_j\) are the corresponding rows of \(V_O^{(l)}\), \(V_R^{(l)}\), and \(V_O^{(l)}\). The GConv layer maps them to intermediate vectors \((\bar v_i,\bar v_r,\bar v_j)\), which are then aggregated into the next-layer features. For every object $o_i \in O$ and edge $(o_i, r, o_j) \in E$, we concatenate the three corresponding sets of feature vectors, which consist of $(v_i, v_r, v_j)$ representing the subject, predicate, and object vectors, respectively. These concatenated feature sets are then passed through the GConv layer for further processing, initiating with an MLP represented as $f_g$, where $(\bar v_i, \bar v_r, \bar v_j) = f_g(v_i, v_r, v_j)$.
This operation computes the set of three intermediate subject, predicate, and object vectors $(\bar{v}_i, \bar{v}_r, \bar{v}_j)$. Subsequently, these intermediate vectors are aggregated to form $V_O^{(l)}$ and $V_R^{(l)}$. Node features $V_O^{(l)}$ include the new feature vectors $v'_i$ for the output nodes and edge features $V_R^{(l)}$ consist of the new feature vectors $v'_r$ for the output predicates:
\begin{align}
    C_i^s & = \{\bar v_i \mid
    (\bar v_i, \bar v_r, \bar v_j) = f_g(v_i, v_r, v_j),\,
    (o_i, r, o_j) \in E\}, \\
    C_i^o & = \{\bar v_i \mid
    (\bar v_j, \bar v_r, \bar v_i) = f_g(v_j, v_r, v_i),\,
    (o_j, r, o_i) \in E\}.
\end{align}
\begin{align}
    v'_i  & = f_o\!\left(v_i,
    \frac{1}{|C_i^s| + |C_i^o|}
    \sum_{\bar v_i \in C_i^s \cup C_i^o} \bar v_i\right), \\
    v'_r  & = f_r(v_r, \bar v_r).
\end{align}
Here, $f_r$ and $f_o$ refer to MLPs that include residual links. Additionally, $C_i^s$ and $C_i^o$ represent candidate vectors for node $o_i$, corresponding to edges where this node serves as either a subject or an object, respectively. To consolidate these candidate vectors for each node, a mean aggregation operation is applied, forming a single vector $v'_i$. On the other hand, the predicate vector $\bar v_r$ is directed through additional neural network layers represented as $f_r$, which generate the new vector $v'_r$ incorporating the effects of the residual link.

Appendix Section~\ref{sec:app_gconv} summarizes the layer-level GConv update used here.

\subsection{Stage 1: Scene Graph Enrichment}

\subsubsection{\textbf{Scene Graph Enricher}}

The \emph{Scene Graph Enricher} \(G_{sg}\) expands an input graph by predicting one added object, its incident edge, and the predicate on that edge. These steps leverage the proposed GCNs along with an \emph{Enriching Edge Detector}.

First, \(G_{sg,obj}\) predicts the added object \(\hat o\). Second, \(G_{sg,edge}\) scores candidate directed edges incident to \(\hat o\). Third, \(G_{sg,pred}\) predicts the predicate \(\hat r\) for the selected edge. Together, these three steps progressively enhance the input scene graph by introducing new objects, edges, and predicates, ultimately enriching the scene's content iteratively.

Our \emph{Scene Graph Enricher} $G_{sg}$ processes the input graph $g$ as
$\hat g = G_{sg}(g)$.
This produces an enriched graph $\hat{g}$. The enriched graph can then be fed back into $G_{sg}$ for another enrichment step, allowing the scene to be expanded iteratively.

$G_{sg, obj}$ is modeled with the proposed GCN. Additional layers and skip connections are introduced to deepen the GCN, avoiding issues like vanishing gradients. $G_{sg, obj}$ results in the creation of an enriching object denoted as $\hat o$, where
$\hat o = G_{sg, obj}(g)$.
This newly generated enriching object is then added to the original graph $g$. The augmented graph, consisting of both the original objects and the newly created object, continues to serve as the input for the succeeding steps of the enrichment process.

\(G_{sg,edge}\) uses two MLPs, \(\phi_s\) and \(\phi_o\), with the same architecture and separate weights. These MLPs receive several inputs: the initial input node feature vectors $V_O^{(0)}$, the enriching object $\hat{o}$, and hidden vectors $V_O^{(L)}$ extracted from the final layer of $G_{sg, obj}$.

The MLPs produce subject-side and object-side representations whose dot products form the edge-score matrix \(\hat M\):
\begin{equation}
    \hat M = \phi_s(V_O^{(0)}, v_{\hat o}, V_O^{(L)}) \cdot \phi_o(V_O^{(0)}, v_{\hat o}, V_O^{(L)})^T
    \label{eq:g_sg_edge}
\end{equation}
Eq.~\ref{eq:g_sg_edge} summarizes this scoring operation; additional architectural details for the \emph{Enriching Edge Detector} are provided in Appendix Section~\ref{sec:app_gcn_impl}.
In the \emph{Enriching Edge Detector} $G_{sg, edge}$, each node vector is transformed into another space to determine the location of the new edges within the enriched graph. From Eq.~\ref{eq:g_sg_edge}, $\hat M$ is the score matrix whose entry $\hat m_{i,j}$ is the dot product between the transformed subject-side representation of node $o_i$ and the transformed object-side representation of node $o_j$. Larger values indicate a higher score for the directed edge from $o_i$ to $o_j$. During training, $\hat M$ is supervised against GT edge existence for candidate pairs involving $\hat o$. During inference, the entries of $\hat M$ incident to the inserted node $\hat o$ define the candidate enriching edges, and the selected enriching edge is
\begin{equation}
    \hat e=\arg\max_{e\in\mathcal{E}(\hat o)} s_{\hat M}(e),
\end{equation}
where $\mathcal{E}(\hat o)$ denotes the candidate directed edges touching $\hat o$, and $s_{\hat M}(e)$ denotes the score read from $\hat M$ for edge $e$. The selected edge has the form $\hat e=(o_e,\cdot,\hat o)$ or $\hat e=(\hat o,\cdot,o_e)$, where $\cdot$ denotes the placeholder predicate to be predicted in the next step.

After selecting an edge \(\hat e\) between \(\hat o\) and an existing object \(o_e\), the resulting graph is passed to \(G_{sg,pred}\). Notably, $G_{sg, pred}$ maintains the same architecture as $G_{sg, obj}$ but operates with an independent set of parameters as
\begin{equation}
    \hat r = G_{sg, pred}(g, \hat o, \hat e).
\end{equation}
$G_{sg, pred}$ is designed with the proposed GCN tasked with producing the type of relation $\hat r$ for each enriching edge $\hat e$ derived from the previous step.

Finally, the new graph $\hat g$ is assembled by combining the input graph $g$, enriching object $\hat o$, the edges $\hat e$, and the relation labels $\hat r$. This comprehensive graph serves as input for the following enrichment iterations and represents the enriched scene with all the appended elements and their associated attributes.

\subsubsection{\textbf{Scene Graph Critic}}

The \emph{Scene Graph Critic} \(D_{sg}\) contains a global discriminator and a local discriminator, both trained jointly with the scene graph enricher.
Additional implementation details for these GCN architectures are provided in Appendix Sections~\ref{sec:app_gconv} and \ref{sec:app_gcn_impl}.
In the global discriminator denoted as $D_{sg, glob}$, a GCN is employed to transform the input into another graph. The extracted features in nodes and edges of this graph are utilized to differentiate the source of input data. Given that our model handles graphs with varying numbers of nodes and edges, an aggregation method such as averaging needs to be utilized at this stage. The output vectors for nodes and edges are separately aggregated and then input into additional neural network layers for further processing and discrimination.

Our global discriminator is deeper compared to the local discriminator represented as $D_{sg, loc}$ by incorporating two additional GConv layers.
The local discriminator receives the subgraph containing the added object \(\hat o\), its immediate neighbors, and the incident edges.
The outputs of two discriminators are concatenated and subsequently passed through additional layers. The concatenated features are mapped to a scalar critic score that estimates whether the input graph is an original graph or an enriched graph.
\subsection{Stage 2: Image Generation}

During training and inference, \(G_{im}\) is a fixed Stable Diffusion image synthesizer \citep{rombach2022high} that renders a text prompt derived from the scene graph.
Both the GT scene graph $g_{GT}$ and the enriched scene graph $\hat g$ are rendered by this same fixed generator to produce images $I$ and $\hat I$ (Eq.~\ref{eq:img_synth}). Let \(\phi(g)\) denote the graph-to-conditioning map used by the renderer. During training, \(\phi\) forms embedding-level conditioning vectors from soft object and predicate distributions. Visual scene and image-text alignment losses are then backpropagated through the fixed image-generation and encoder components to update the trainable scene graph enrichment modules, while the Stable Diffusion, scene-classifier, and CLIP weights remain fixed. The rendered images are used by the visual scene loss in Eq.~\ref{eq:vis_eval} and by the image-text alignment loss in Eq.~\ref{eq:im_text_clid}.
\begin{equation}
    I = G_{im}(\phi(g_{GT})), \;\;\; \hat I = G_{im}(\phi(\hat g)).
    \label{eq:img_synth}
\end{equation}

\subsection{Stage 3: Training-Time Visual Alignment}

\subsubsection{\textbf{Visual Scene Characterizer}}
The visual scene characterizer encourages the rendered enriched image \(\hat I\) to preserve the coarse scene context of the GT-rendered image \(I\). Thus, a pre-trained scene classifier $S_{cf}$ \citep{zhou2017places} receives the generated enriched image $\hat{I}$ to evaluate the mentioned criteria. The hidden scene features for both images $I$ and $\hat I$ are extracted (Eq.~\ref{eq:vis_eval}). These extracted features are afterward integrated into the objective function, thus playing a crucial role in the joint training objective.
\begin{equation}
    h_I = S_{cf}(I), \;\;\; h_{\hat I} = S_{cf}(\hat I).
    \label{eq:vis_eval}
\end{equation}

\subsubsection{\textbf{Image-Text Aligner}}

The \emph{Image-Text Aligner} \(M_{im\_sg}\) uses pretrained CLIP encoders \citep{radford2021learning}. It encourages the enriched generated images to remain aligned with the essential scene elements in the input description. This multimodal network operates by accepting both the input graph $g$ and the enriched image $\hat I$. It extracts features denoted as $F_g$ from the graph and $F_I$ from the image:
\begin{equation}
    F_g = M_{im\_sg}(g), \;\;\; F_I = M_{im\_sg}(\hat I).
\end{equation}
These features are further employed in the objective function.
\subsection{Objective Function}

For supervised graph-enrichment training, we remove one object and its incident edges from each GT scene graph \(g_{GT}\). The remaining graph is the input graph \(g\), and the removed object and relations provide supervision.
The overall objective combines graph-completion supervision, alignment terms, and adversarial supervision.
The goal of $G_{sg, obj}$ is to predict the object category of this removed node. Since $\hat o$ is selected from the discrete object vocabulary, this leads to our cross-entropy \emph{Obj. Loss},
\begin{equation}
    \LL_{obj} = \frac{1}{N}\sum_{i=1}^N CE(\hat o^{(i)}, o_{GT}^{(i)}),
\end{equation}
where $N$ represents the mini-batch size. This loss is structured as a cross-entropy measure, comparing the predicted object $\hat o^{(i)}$ and its GT counterpart $o_{GT}^{(i)}$ in the GT graph $g_{GT}$. During training, the removed object is represented as a temporary object slot, and all candidate edges incident to this slot are scored before predicate prediction.

\(G_{sg,edge}\) produces an edge-score matrix \(\hat M\), where each entry \(\hat m_{i,j}\in[0,1]\) estimates the probability of a directed edge from \(o_i\) to \(o_j\). The \emph{Enriching Edge Detector} contributes to the objective function with the binary cross-entropy \emph{Edge Loss},
\begin{equation}
    \LL_{edge} = \frac{1}{N}\sum_{k=1}^N \sum_{i, j} BCE(\hat m_{i, j}^{(k)}, m^{(k)}_{GT; i, j})
\end{equation}
computed as a binary cross-entropy considering whether an edge exists between each pair of nodes in the GT graph, including the enriching object.

Based on $\hat M$, one enriching edge is selected and its predicate is then predicted by $G_{sg, pred}$. Because this predicate prediction is categorical, we use two cross-entropy terms: \emph{Available Predicate Loss} when the selected GT edge has an annotated predicate, and \emph{No-Relation Predicate Loss} when the selected case is treated as a no-relation target:
\begin{align}
    \LL_{pr, a}  & =
    \frac{1}{N}\sum_{i=1}^N CE(\hat r^{(i)}, r_{GT}^{(i)}), \\
    \LL_{pr, na} & =
    \frac{1}{N}\sum_{i=1}^N CE(\hat r^{(i)}, r_{\emptyset}),
\end{align}
where \(r_{\emptyset}\) denotes the no-relation predicate category used when no annotated predicate is available.
Using the features from Eq.~\ref{eq:vis_eval}, the \emph{Scene Classifier Loss} compares the GT-rendered image and enriched image with an \(L_1\) distance:
\begin{equation}
    \LL_{sc} = \frac{1}{N}\sum_{i=1}^N \vert\vert h_I^{(i)} - h_{\hat I}^{(i)}\vert\vert_1.
\end{equation}
This feature distance encourages the rendered enriched image to preserve the coarse scene characteristics of the GT-rendered image.

The \emph{Image-Text Aligner} produces normalized graph and image features, \(F_g\) and \(F_I\), which define the \emph{Image-SG Alignment Loss}:
\begin{equation}
    \LL_{im\_sg} = - \frac{1}{N}\sum_{i=1}^N F_g^{(i)} \cdot F_I^{(i)},
    \label{eq:im_text_clid}
\end{equation}
which takes the form of a cosine similarity because the CLIP-based features are normalized. Appendix Section~\ref{sec:app_clip_text} gives the graph-to-text linearization used for CLIP text encoding.
This similarity measure is applied to maintain consistency between the synthesized enriched image and the fundamental scene semantics conveyed by the original description through the input graph.

$D_{sg}$ and $G_{sg}$ undergo training within an adversarial setting by engaging in a GAN min-max game. In this setting, the generator and the discriminators have opposing objectives, seeking to minimize and maximize the following loss term, respectively. Let \(p_{\textrm{enriched}}\) denote the distribution of enriched graphs \(\hat g=G_{sg}(g)\) produced by the scene graph enricher from input graphs \(g\):
\begin{equation}
    \begin{aligned}
        \LL_{GAN} ={} & \EE_{g\sim p_{\textrm{real}}} \log D_{sg}(g)                        \\
                      & + \EE_{\hat g \sim p_{\textrm{enriched}}} \log(1 - D_{sg}(\hat g)).
    \end{aligned}
\end{equation}

Finally, in our joint graph-enrichment training process, we use a weighted sum $\LL$ that encompasses all the previously mentioned loss terms (Eq.~\ref{eq:w_1oss}). The weights balance losses with different numerical scales and training roles, so object, edge, predicate, adversarial, scene-consistency, and image-text alignment signals can contribute during joint training.
\begin{equation}
    \begin{aligned}
        \LL ={} & w_0\LL_{obj}
        + w_1\LL_{edge}
        + w_2\LL_{GAN}
        + w_3\LL_{pr, a}            \\
                & + w_4\LL_{pr, na}
        + w_5\LL_{sc}
        + w_6\LL_{im\_sg}
    \end{aligned}
    \label{eq:w_1oss}
\end{equation}

Training hyperparameters, the selected configuration, and loss-weight choices are summarized in Appendix Section~\ref{sec:app_training_details}; discriminator-capacity and update-frequency choices are detailed in Appendix Section~\ref{sec:app_pair_discriminators}. These appendix sections cover our GCN architectures, mini-batch size, activation functions, embedding dimensions, dropout, normalization, the selected loss weights, and discriminator-capacity choices.

During training, gradients from the visual scene and image-text alignment losses pass through the frozen conditioning, image-generation, scene-classification, and CLIP components to the trainable scene graph modules; only \(G_{sg}\) and \(D_{sg}\) are updated.

During inference, the framework inserts a temporary object slot and scores candidate incident edges before applying $G_{sg}$. The edge with the largest score incident to the inserted slot is selected, and the resulting object and predicate predictions determine the next enriched graph state. For a fixed trained enrichment module and fixed edge-selection rule, the graph-enrichment stage is deterministic. Different enrichment variants can still be obtained by choosing a different ranked edge at any iteration. By contrast, the graph-to-image rendering stage is stochastic and can yield different images from the same graph-derived text. The enrichment process can be iteratively performed multiple times to progressively add additional content. In the graph-side proxy metrics, one enrichment step predicts the held-out object and its incident relation. For image-quality metrics, qualitative examples, and user-study stimuli, we apply iterative enrichment with the same edge-selection rule and keep the final graph-derived text within the 25-word rendering protocol so that the fixed image generator is less likely to ignore added semantics in overlong prompts.
\section{Experiments}

We evaluate whether the proposed GCE framework predicts coherent graph enrichments and whether those enrichments improve rendered images while preserving the input scene. We assess these goals through qualitative illustrations, graph-side proxy metrics, image quality metrics, ablations, and user studies.

We also include a comparison with ChatGPT as an LLM. The contribution here is the overall framework for GCE rather than a claim that the current GCN, scene classifier, or renderer is individually stronger than specialized alternatives for their own tasks. Modules with similar roles could be substituted within the same high-level framework without changing the task definition.

Concretely, this section reports dataset and setup choices, a text-only ChatGPT baseline family, quantitative results for proxy scene graph enrichment and image quality metrics, ablations, and user studies.

For image-level visualization and metrics, each simple scene representation (directly rendered from the sparse input without graph enrichment) or enriched scene representation is converted to text and rendered by the fixed Stable Diffusion image synthesizer \citep{rombach2022high}. We use this same renderer for the reported image comparisons so that differences are attributed to the input description or enrichment method rather than to different image generators.

\subsection{Dataset}

We use the Visual Genome (VG) dataset \citep{krishna2017visual}, which consists of 110k real-world images with their scene graphs. We follow preprocessing similar to \cite{johnson2018image}. This resulted in 178 object and 45 predicate categories in approximately 62.5k training samples, 5.5k validation samples, and 5k test samples. VG provides rich scene annotations, including 3.8M objects and 2.3M relations across diverse environments and object configurations.

We focus on Visual Genome because GCE evaluation requires human-annotated object--predicate--object scene graphs. Other image datasets with object or region labels do not provide the same full object--predicate--object scene-graph supervision required for this evaluation; broader cross-dataset evaluation remains future work.

\subsection{Proxy Scene Graph Enrichment Metrics}
\label{sec:proxy_scene_graph_metrics}

Because a sparse scene can have multiple valid enrichments, the graph metrics below are proxy recovery and plausibility measures rather than a complete definition of GCE quality. We therefore report proxy scene graph enrichment metrics that test whether a method can recover annotated, semantically plausible enrichments under controlled reductions of Visual Genome scene graphs. In the single-object removal protocol, one object is sampled from each test scene graph and is removed together with all incident relations. The evaluated method receives the reduced graph and predicts the held-out object and its incident structure. To avoid depending on one arbitrary removed object, the final comparison repeats the reduction with multiple random object-removal seeds and reports mean \(\pm\) standard deviation for each metric. The Visual Genome split, trained checkpoints, and metric definitions remain fixed; only the randomly removed object in each graph changes across seeds. Object accuracy is computed over all 5k test cases. Triplet, edge, predicate, and MEP+ metrics are computed over the visible-target cases, where the held-out object has at least one ground truth incident triplet after filtering. Empty-target cases are retained only for object scoring, so object recovery is not biased by dropping cases without visible incident triplets.

We compare our scene graph enricher with SceneGraphGen+, our GCE adaptation of SceneGraphGen \citep{scenegraphgen2021}. SceneGraphGen is an unconditional scene graph generation method rather than a GCE method. SceneGraphGen+ applies its graph-completion interface to the same reduced graphs. For evaluation, we convert its completed graph into the one-step GCE output format: one predicted object and the incident triplets attached to that object. This gives a practical external graph generation baseline for GCE. Other nearby scene graph methods are less directly comparable because they usually infer graphs from images, render images from specified graphs, or require a conditioning interface that does not expose the same seed graph to enriched graph mapping. A stronger learned graph-completion baseline would also need a dedicated training and evaluation interface for mapping reduced graphs to enriched graphs under the same Visual Genome reduction protocol; we leave that broader baseline study to future work.

All graph-side learned methods are evaluated on variable-size reduced scene graphs. Because the GCN-based enricher uses shared relation-triplet update functions, the same learned parameters can process reduced graphs with different numbers of objects and edges.

We also evaluate simple graph-side controls with the same reduced-graph protocol. \emph{Random prior} samples the added object and incident relation pattern from training-set label frequencies. \emph{Global frequency prior} predicts the most frequent object and relation pattern observed in the training split. \emph{Conditional co-occurrence prior} conditions on labels present in the reduced seed graph and selects frequent co-occurring added objects and predicates. \emph{Nearest-neighbor retrieval} retrieves a similar training graph context and reuses its removed-object enrichment pattern. These controls test whether learned graph enrichment improves over simple dataset-prior and retrieval behavior.

Table~\ref{tab:expanded_setting_a_canonical_full} reports the main single-object removal comparison between simple controls, SceneGraphGen+, and our method. The main columns summarize held-out object recovery, incident-edge recovery, predicate recovery, triplet recovery, and the auxiliary MEP+ support score. Triplet precision \(P_{\mathrm{tri}}\), recall \(R_{\mathrm{tri}}\), and F1 \(F_{\mathrm{tri}}\) evaluate recovered incident triplets. The triplet entries are percentage-point displays of underlying fractions: a displayed value \(x\) denotes an underlying fraction \(x/100\). The reported triplet F1 is a macro average of per-case F1 values, so it is not recomputed from the averaged precision and recall. MEP+ is a context-support precision metric inspired by modified edge precision in GEMS \citep{agarwal2023gems}; it measures whether predicted added triplets are supported by training graphs with the same reduced context, and higher values are better. MEP+ and no-edge accuracy are auxiliary diagnostics rather than primary evidence of semantic quality: MEP+ is strict about training-context support, and no-edge accuracy can be inflated by sparse annotations. Appendix Section~\ref{sec:app_expanded_proxy_metrics} gives formal definitions for these metrics in Eqs.~\ref{eq:metric_object_accuracy}--\ref{eq:metric_mep_plus}.

These graph-output metrics should be interpreted as proxy evidence rather than a complete measure of GCE quality, because multiple enrichments can be plausible for the same sparse scene. The table therefore emphasizes controlled recovery and plausibility under the same reduced graph protocol, while the qualitative and image-level evaluations assess complementary aspects of the generated content. Improvements in this table provide relative evidence within the controlled Visual Genome protocol, and the broader GCE behavior is assessed through complementary qualitative examples and user studies.
Additional selected-configuration and hyperparameter details are summarized in Appendix Section~\ref{sec:app_training_details}.

\begin{table*}
    \centering
    \vspace{\tabup}
    \caption{\emph{Proxy Scene Graph Enrichment Metrics on Visual Genome (single-object removal).} Entries report display values \(\pm\) standard deviations over repeated random object-removal seeds. Object accuracy uses all 5k test cases. Available-edge accuracy is recall over ground-truth incident edge slots; available-predicate accuracy is conditional on slots active in both the prediction and GT; triplet and MEP+ metrics use visible-target cases.}
    \label{tab:expanded_setting_a_canonical_full}
    \vspace{\tabmid}
    \scriptsize
    \setlength{\tabcolsep}{2.6pt}
    \resizebox{\textwidth}{!}{%
        \begin{tabular}{@{}lcccccccc@{}}
            \toprule
            \textbf{Method}       & \textbf{$A_{\mathrm{obj}}$} & \textbf{$A_{\mathrm{edge}}$} & \textbf{$A_{\mathrm{noedge}}$} & \textbf{$A_{\mathrm{pred}}$} & \textbf{$P_{\mathrm{tri}}$} & \textbf{$R_{\mathrm{tri}}$} & \textbf{$F_{\mathrm{tri}}$} & \textbf{MEP+}   \\ \midrule
            Random                & 2.05 $\pm$ 0.20             & 1.01 $\pm$ 0.20              & 99.27 $\pm$ 0.06               & 0.25 $\pm$ 0.09              & 0.01 $\pm$ 0.01             & 0.00 $\pm$ 0.01             & 0.00 $\pm$ 0.01             & 0.43 $\pm$ 1.36 \\
            Global freq.          & 2.74 $\pm$ 0.26             & 16.13 $\pm$ 1.10             & 90.44 $\pm$ 0.13               & 4.31 $\pm$ 0.39              & 0.03 $\pm$ 0.04             & 0.03 $\pm$ 0.03             & 0.03 $\pm$ 0.03             & 1.76 $\pm$ 2.27 \\
            Cond. co-occur.       & 6.12 $\pm$ 0.34             & 8.31 $\pm$ 0.42              & 95.45 $\pm$ 0.26               & 2.75 $\pm$ 0.27              & 0.69 $\pm$ 0.13             & 0.67 $\pm$ 0.15             & 0.63 $\pm$ 0.12             & 2.63 $\pm$ 1.08 \\
            Nearest-neighbor      & 6.19 $\pm$ 0.32             & 9.97 $\pm$ 1.19              & 94.67 $\pm$ 0.33               & 4.02 $\pm$ 0.53              & 0.98 $\pm$ 0.10             & 1.09 $\pm$ 0.11             & 0.97 $\pm$ 0.10             & 2.08 $\pm$ 0.74 \\
            SceneGraphGen+        & 15.19 $\pm$ 3.68            & 36.64 $\pm$ 3.16             & 92.41 $\pm$ 0.61               & 14.17 $\pm$ 1.18             & 4.87 $\pm$ 0.52             & 6.24 $\pm$ 1.04             & 3.73 $\pm$ 0.53             & 5.56 $\pm$ 4.18 \\
            Ours                  & 20.60 $\pm$ 2.21            & 75.18 $\pm$ 5.62             & 98.90 $\pm$ 0.54               & 42.20 $\pm$ 5.94             & 31.69 $\pm$ 5.34            & 12.30 $\pm$ 1.99            & 7.73 $\pm$ 1.24             & 7.19 $\pm$ 2.54 \\ \bottomrule
        \end{tabular}%
    }
    \vspace{\tabdwn}
\end{table*}

Across 10 repeated object-removal seeds, our method gives the strongest primary recovery scores: it has the highest object accuracy, available-edge accuracy, available-predicate accuracy, and triplet precision, recall, and F1. Compared with SceneGraphGen+, it improves available-edge accuracy from \(36.64\) to \(75.18\), available-predicate accuracy from \(14.17\) to \(42.20\), and triplet precision from \(4.87\) to \(31.69\). Its triplet macro-F1 score \(F_{\mathrm{tri}}\) also remains higher, increasing from \(3.73\) for SceneGraphGen+ to \(7.73\) for our method, while the auxiliary MEP+ score increases from \(5.56\) to \(7.19\). The simple prior and nearest-neighbor controls remain substantially lower. The no-edge and MEP+ columns should be read as auxiliary diagnostics: no-edge accuracy is dominated by sparse annotations, and MEP+ remains small under the strict same-context support criterion.

\subsection{Text-Based Baselines and Image Quality Metrics}

As a text-space baseline, we compare against ChatGPT prompt expansion \citep{openai2023gpt4}.  In this paper, we use ChatGPT prompt expansion as a text-based comparison point rather than as a stand-in for all specialized LLM systems. First, jointly training LLMs and image generators would require substantial data and computational resources. Second, although LLMs can enrich text, they do not explicitly evaluate whether the added content remains visually grounded and suitable for one coherent generated image. Third, out-of-the-box LLM text expansion can produce redundant content for image synthesis, particularly because even strong image-generation methods \citep{rombach2022high} can struggle with overly complex prompts.

For the image quality comparison, we evaluate direct text-to-image generation from the input description (\emph{Simple}) and three ChatGPT prompt-enrichment variants under the same downstream image synthesizer. Within this local comparison setting, the proposed framework outperforms the ChatGPT-based baselines. This comparison isolates whether explicit scene graph enrichment provides an empirical advantage over LLM-based text expansion under the same downstream image-synthesis pipeline. This ChatGPT baseline is a relevant out-of-the-box text-space baseline for GCE, whereas our method instantiates GCE through explicit scene graph enrichment. Fig.~\ref{fig:samples} compares \emph{Simple}, \emph{ChatGPT Direct}, and \emph{Ours} qualitatively, while Table~\ref{tab:fid_is} compares \emph{Simple}, the three ChatGPT prompt variants, \emph{Enriched Ours}, \emph{GT Synthesized}, and real \emph{VG} images. The input descriptions are provided to ChatGPT, which generates enriched descriptions that are then used to synthesize images.

We limited ChatGPT outputs to 15--25 words because long prompt expansions can be difficult for image generators to render coherently. We use the same 25-word rendering protocol for graph-derived texts in the reported image comparisons. We evaluate three ChatGPT prompt-expansion styles: direct completion, scene-focused completion, and object-focused completion. These baselines test a simple text-space alternative under the same renderer; they are not intended to represent the strongest possible LLM system or optimized LLM prompt-engineering setup. The exact prompt templates are provided in Appendix Section~\ref{sec:app_chatgpt_prompts}.
\subsubsection{\textbf{Image Quality Metrics}}

\begin{table*}
    \centering
    \vspace{\tabup}
    \caption{\emph{Generated Image Quality Metrics.} Inception Scores (IS) \citep{salimans2016improved} and Fréchet Inception Distances (FID) \citep{heusel2017gans} are reported for real \emph{VG} images and for image sets rendered from GT scene descriptions (\emph{GT Synthesized}), simple descriptions (\emph{Simple}), and enriched descriptions produced by our method and the three ChatGPT prompt variants. All generated images use the same final Stable Diffusion rendering pipeline.}
    \label{tab:fid_is}
    \vspace{\tabmid}
    \resizebox{\textwidth}{!}{%
        \begin{tabular}{@{}lcccccccc@{}}
            \toprule
            \textbf{Metric}                                                                                       &
            \textbf{Simple}                                                                                       &
            \textbf{GPT Direct}                                                                                   &
            \textbf{GPT Scene}                                                                                    &
            \textbf{GPT Object}                                                                                   &
            \textbf{Ours}                                                                                         &
            \textbf{\textit{GT Synth.}}                                                                           &
            \textbf{\textit{VG}}                                                                                  &
            \textbf{\begin{tabular}[c]{@{}c@{}}Ours vs.\\ Simple\end{tabular}}                                      \\ \midrule
            \textbf{IS$\uparrow$}                                                                                 &
            \multicolumn{1}{c}{17.03 $\pm$ 0.37}                                                                  &
            \multicolumn{1}{c}{15.44 $\pm$ 0.32}                                                                  &
            \multicolumn{1}{c}{14.90 $\pm$ 0.19}                                                                  &
            \multicolumn{1}{c}{15.69 $\pm$ 0.26}                                                                  &
            \multicolumn{1}{c}{\textbf{19.37 $\pm$ 0.28}}                                                         &
            \multicolumn{1}{c}{\textit{24.44 $\pm$ 0.32}}                                                         &
            \multicolumn{1}{c}{\textit{38.61 $\pm$ 0.52}}                                                         &
            13.74\%                                                                                                 \\
            \textbf{FID$\downarrow$ [GT Synth.]}                                                                  &
            \multicolumn{1}{c}{37.85}                                                                             &
            \multicolumn{1}{c}{42.15}                                                                             &
            \multicolumn{1}{c}{47.74}                                                                             &
            \multicolumn{1}{c}{52.74}                                                                             &
            \multicolumn{1}{c}{\textbf{34.89}}                                                                    &
            \multicolumn{1}{c}{\textit{0}}                                                                        &
            \multicolumn{1}{c}{\textit{35.32}}                                                                    &
            7.82\%                                                                                                  \\
            \textbf{FID$\downarrow$ [VG]}                                                                         &
            61.27                                                                                                 &
            64.69                                                                                                 &
            68.06                                                                                                 &
            73.48                                                                                                 &
            \textbf{53.23}                                                                                        &
            \textit{35.32}                                                                                        &
            \textit{0}                                                                                            &
            13.12\%                                                                                                 \\ \bottomrule
        \end{tabular}%
    }
    \vspace{\tabdwn}
\end{table*}

Inception Scores (IS) \citep{salimans2016improved} and Fréchet Inception Distances (FID) \citep{heusel2017gans} are widely recognized metrics used to evaluate the quality of synthesized images. IS relies on a pre-trained image classifier to assess the presence of recognizable objects and their diversity within the generated image set, and we report it in the usual mean $\pm$ standard deviation form. FID uses a pre-trained classifier to extract deep features from generated and real image sets. It then measures the statistical similarity between the distributions of these extracted features for the generated and real image sets, so each reported FID value is a single statistic for that pair of image sets. IS and FID capture useful distribution-level properties of the generated images, but they do not directly measure whether a specific added object or relation is semantically correct.

Table~\ref{tab:fid_is} reports FID and IS for evaluating image quality. Here, \emph{Simple} denotes direct rendering from the input description without enrichment, \emph{Enriched ChatGPT Direct/Scene/Object} denote the three text-only prompt-expansion variants, and \emph{Enriched Ours} denotes our graph-based enrichment pipeline. We use Stable Diffusion as the shared image synthesizer and compare the generated images with real Visual Genome images:

The \emph{Simple} image set contains 36.5k images rendered directly from test-set input descriptions without explicit enrichment.
The four \emph{Enriched} image sets use the same input descriptions after enrichment by ChatGPT Direct, ChatGPT Scene, ChatGPT Object, or our graph-based framework.
The \emph{GT Synthesized} image set contains 35k images rendered from GT training scene descriptions.
The \emph{VG} image set contains 87k real Visual Genome images.
The Simple and Enriched image sets have matched sample counts within each FID comparison. The GT Synthesized and VG columns are reference anchors with the stated sample sizes, so FID values should be interpreted within the same reference row rather than compared across different reference distributions.

For FID, we use two reference distributions: \emph{GT Synthesized} and real \emph{VG} images. Thus, \emph{FID [GT Synthesized]} uses \emph{GT Synthesized} as the reference distribution, whereas \emph{FID [VG]} uses \emph{VG} as the reference distribution. The \emph{GT Synthesized} reference helps compare methods under the same rendering pipeline and reduces the influence of the domain gap between generated and real images; the \emph{VG} reference measures distance to real-image statistics.
The last column reports the relative gain for IS and the relative reduction for FID when comparing \emph{Enriched Ours} with \emph{Simple}.
Relative to \emph{Simple}, our graph-based enriched variant improves the reported IS and FID values in Table~\ref{tab:fid_is}. These scores suggest improved image-set quality under the fixed renderer, but they do not directly verify whether the added objects and relations are semantically correct; for that reason, we read them together with the graph-side proxy metrics, qualitative examples, and user studies rather than as standalone proof of GCE quality. \emph{GT Synthesized} remains the strongest generated-reference column because it is rendered from GT training descriptions, which are more detailed and diverse than reduced test descriptions. Under this fixed image-synthesis setup, \emph{GT Synthesized} serves as a strong comparison anchor rather than a task-level upper bound.

Due to limitations in the Visual Genome dataset, our model does not enrich the attributed properties of each object, such as its color. However, these attributes are present in ChatGPT-enriched texts, which adds diversity to the generated images, potentially improving IS and FID. As a result, the ChatGPT prompt-expansion variants may have an a priori advantage on IS and FID because they can add attribute words that our graph annotations do not supply. Within this reported setup, our graph-based enriched variant outperforms the three ChatGPT prompt variants on the reported FID and IS comparisons, while the ChatGPT variants remain below \textit{Simple} on these metrics. In this out-of-the-box prompt-expansion setting, moving from \textit{ChatGPT Direct} to \textit{Scene} to \textit{Object} adds progressively more free-form text, yet the reported FID values worsen rather than improve. One plausible explanation is that longer text-only expansions can add visually redundant or weakly grounded details without controlling how those details compose into one coherent image.

\subsection{Ablation Study}

To probe the effect of major components, we compare the final configuration with variants that remove one module or alter the discriminator structure. The results of this study are presented in Table \ref{tab:ablation},  where the ablated models are compared employing our defined metrics for scene graph enrichment. These ablations assess how each component affects graph-side proxy recovery metrics. Since this task lacks a unique solution, raw results with the specified metrics should be read comparatively and cautiously; they indicate how each module affects the current proxy measures, but they do not fully isolate each module's contribution to the overall GCE task. The differences across ablations are modest in these proxy metrics, which is consistent with the modules contributing jointly through the selected framework.

\begin{table}
    \centering
    \vspace{\tabup}
    \caption{\emph{Ablation Study Results.} Each variant changes one module relative to the final configuration and is evaluated with the same proxy scene graph enrichment metrics used throughout the experiments.}
    \label{tab:ablation}
    \vspace{\tabmid}
    \resizebox{0.9\columnwidth}{!}{%
        \begin{tabular}{ccccccc}
            \toprule
            \textbf{Measure}                                      & \textbf{w/o $M_{im\_sg}$} & \textbf{W/o $D_{sg}$} & \textbf{W/o $D_{sg, glob}$} & \textbf{W/o $D_{sg, loc}$} & \textbf{W/o $S_{cf}$} & \textbf{Ours} \\ \midrule
            \textbf{Object Acc. ($A_{\mathrm{obj}}$)}             & 19.04                     & 18.33                 & 18.52                       & 18.66                      & 17.88                 & 20.60         \\ \midrule
            \textbf{Avail. Edge Acc. ($A_{\mathrm{edge}}$)}       & 75.00                     & 74.89                 & 74.20                       & 75.33                      & 74.62                 & 75.18         \\
            \textbf{No-Edge Acc. ($A_{\mathrm{noedge}}$)}         & 99.58                     & 99.61                 & 99.61                       & 99.60                      & 99.66                 & 98.90         \\
            \textbf{Avail. Pred. Acc. ($A_{\mathrm{pred}}$)}      & 41.95                     & 37.43                 & 38.87                       & 37.47                      & 36.08                 & 42.20         \\ \bottomrule
        \end{tabular}%
    }
    \vspace{\tabdwn}
\end{table}

The reference configuration for these studies is the final model, \emph{Ours}. Its selected hyperparameters and architectures are summarized in Appendix Section~\ref{sec:app_training_details}. We explore variants that remove the image-text aligner, remove the scene classifier, remove the pair of discriminators, or remove either the local or global discriminator individually.

In the case labeled \emph{\textbf{W/o $D_{sg}$}}, we omit the pair of discriminators and therefore remove the associated adversarial training signal.

\emph{\textbf{W/o $D_{sg, loc}$}} and \emph{\textbf{W/o $D_{sg, glob}$}} are cases with only global or local discriminators retained, respectively. \emph{W/o $D_{sg, loc}$} may place relatively more weight on whole-scene signals than on local enriched regions, whereas \emph{W/o $D_{sg, glob}$} may place relatively more weight on local enriched regions than on whole-scene consistency.

\emph{\textbf{W/o $S_{cf}$}} omits the pre-trained scene classifier in the \emph{Visual Scene Characterizer}. This component evaluates whether the synthesized images reflect the essential scene characteristics described in the original input. Although some synthesized objects are difficult for the scene classifier to recognize, the associated loss still changes the reported proxy metrics and is therefore retained in the baseline.
The variant labeled \emph{\textbf{w/o $M_{im\_sg}$}} removes the \emph{Image-Text Aligner}, a CLIP-based component that encourages enriched images to remain aligned with the original scene graph semantics.
\subsection{Qualitative Results}

\newlength{\imgwidth}
\setlength{\imgwidth}{0.15\textwidth}
\newlength{\txtwidth}
\setlength{\txtwidth}{0.15\textwidth}
\begin{figure*}
    \vspace{\figup}
    \centering
    \hspace{3mm}\begin{minipage}{\txtwidth} \centering\scriptsize (a)\\Input Description \end{minipage}\hspace{0.5mm}
    \begin{minipage}{\txtwidth} \centering\scriptsize (b)\\Enriched Description\\ChatGPT \end{minipage}\hspace{0.5mm}
    \begin{minipage}{\txtwidth} \centering\scriptsize (c)\\Enriched Description\\Ours \end{minipage}\hspace{1.0mm}
    \begin{minipage}{\imgwidth} \centering\scriptsize (d)\\Simple Image\\(Directly by Input Description) \end{minipage}\hspace{-0.3mm}
    \begin{minipage}{\imgwidth} \centering\scriptsize (e)\\Enriched Image\\ChatGPT \end{minipage}\hspace{-0.3mm}
    \begin{minipage}{\imgwidth} \centering\scriptsize (f)\\Enriched Image\\Ours \end{minipage}
    \\
    \vspace{1mm}
    \raisebox{0.5\imgwidth}{
        \begin{minipage}[c]{\txtwidth}
            \centering \footnotesize
            Railing around Edge. Water.
        \end{minipage}}\hspace{-0.5mm}
    \raisebox{0.5\imgwidth}{
        \begin{minipage}[c]{\txtwidth}
            \centering \footnotesize
            Railing around edge. Water below. A tranquil view, with safety in mind, encapsulates the beauty of this waterfront setting.
        \end{minipage}}\hspace{-0.5mm}
    \raisebox{0.5\imgwidth}{
        \begin{minipage}[c]{\txtwidth}
            \centering \footnotesize
            Railing \textbf{on sidewalk} and around edge. Water. \textbf{ Sky has} edge. \textbf{Tree on sidewalk that has beach, ocean, grass, and cloud.}
        \end{minipage}}\hspace{1.0mm}
    \includegraphics[width=\imgwidth]{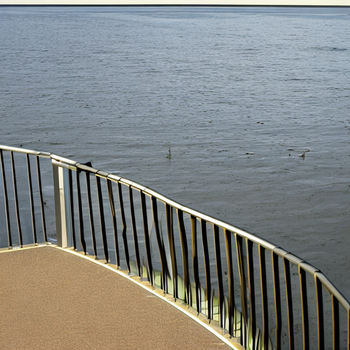}\hspace{-0.5mm}
    \includegraphics[width=\imgwidth]{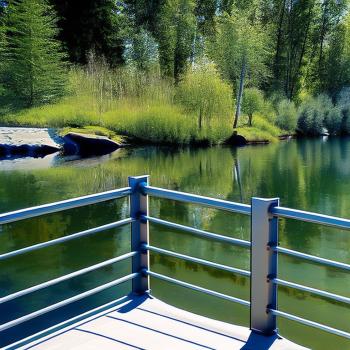}\hspace{-0.5mm}
    \includegraphics[width=\imgwidth]{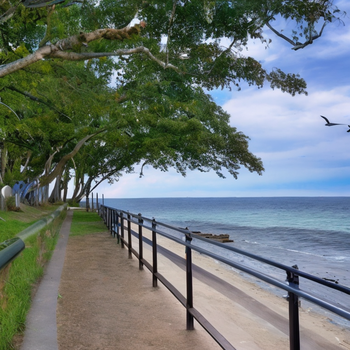}\\
    \vspace{1mm}
    \raisebox{0.5\imgwidth}{
        \begin{minipage}[c]{\txtwidth}
            \centering \footnotesize
            Plate next to bowl. Glass next to bowl
        \end{minipage}}\hspace{-0.5mm}
    \raisebox{0.5\imgwidth}{
        \begin{minipage}[c]{\txtwidth}
            \centering \footnotesize
            Plate and glass next to bowl. The table set for a simple, yet inviting meal, ready to be enjoyed.
        \end{minipage}}\hspace{-0.5mm}
    \raisebox{0.5\imgwidth}{
        \begin{minipage}[c]{\txtwidth}
            \centering \footnotesize
            Plate and glass next to bowl \textbf{on top of table. Food, cup, leaf, flower, water on top of} plate.
        \end{minipage}}\hspace{1.0mm}
    \includegraphics[width=\imgwidth]{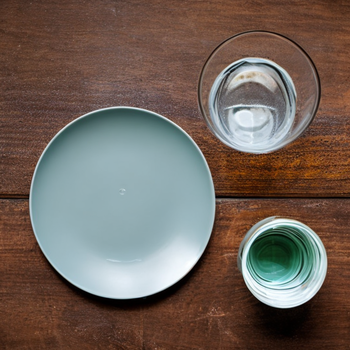}\hspace{-0.5mm}
    \includegraphics[width=\imgwidth]{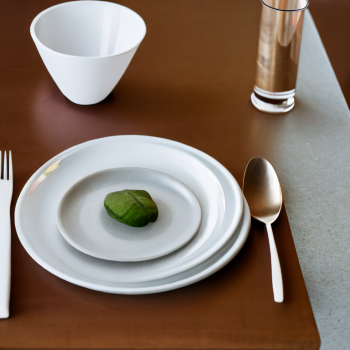}\hspace{-0.5mm}
    \includegraphics[width=\imgwidth]{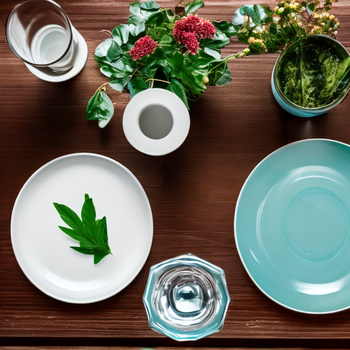}\\
    \vspace{1mm}
    \raisebox{0.5\imgwidth}{
        \begin{minipage}[c]{\txtwidth}
            \centering \footnotesize
            Cloud above water. Reflection on top of water.
        \end{minipage}}\hspace{-0.5mm}
    \raisebox{0.5\imgwidth}{
        \begin{minipage}[c]{\txtwidth}
            \centering \footnotesize
            Cloud above water. Reflection on top of water. The peaceful lake mirrors the sky, creating a serene and beautiful sight.
        \end{minipage}}\hspace{-0.5mm}
    \raisebox{0.5\imgwidth}{
        \begin{minipage}[c]{\txtwidth}
            \centering \footnotesize
            Cloud \textbf{in mountain.} Cloud above water. Reflection on top of water. \textbf{Sky, tree, building, boat, and house in} water.
        \end{minipage}}\hspace{1.0mm}
    \includegraphics[width=\imgwidth]{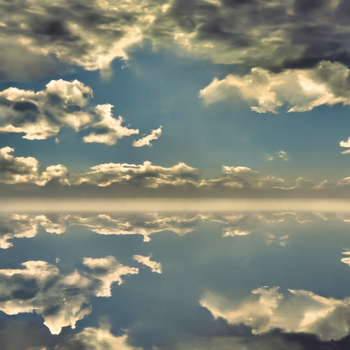}\hspace{-0.5mm}
    \includegraphics[width=\imgwidth]{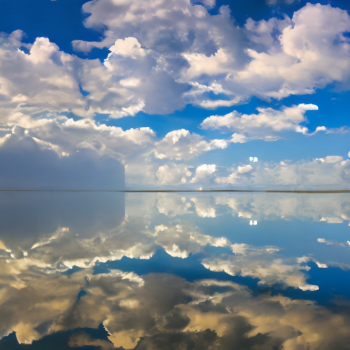}\hspace{-0.5mm}
    \includegraphics[width=\imgwidth]{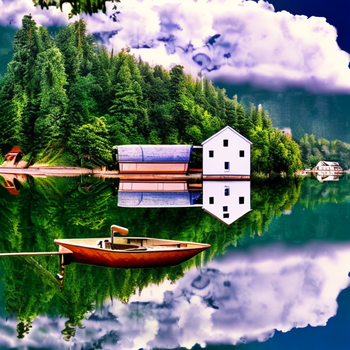}\\
    \vspace{1mm}
    \raisebox{0.5\imgwidth}{
        \begin{minipage}[c]{\txtwidth}
            \centering \footnotesize
            Glass with water. Plate with pizza.
        \end{minipage}}\hspace{-0.5mm}
    \raisebox{0.5\imgwidth}{
        \begin{minipage}[c]{\txtwidth}
            \centering \footnotesize
            Glass with water. Plate with pizza. A delicious meal, with a refreshing drink, awaits, satisfying both thirst and appetite.
        \end{minipage}}\hspace{-0.5mm}
    \raisebox{0.5\imgwidth}{
        \begin{minipage}[c]{\txtwidth}
            \centering \footnotesize
            Glass with water \textbf{on table.} Plate with pizza. \textbf{Cup, handle, bottle, food, bowl, glasses, and hand on table.}
        \end{minipage}}\hspace{1.0mm}
    \includegraphics[width=\imgwidth]{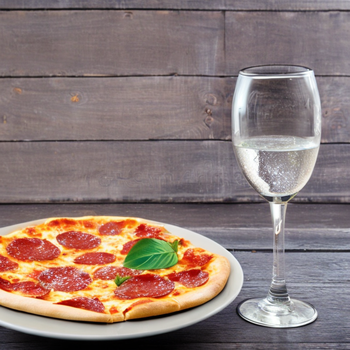}\hspace{-0.5mm}
    \includegraphics[width=\imgwidth]{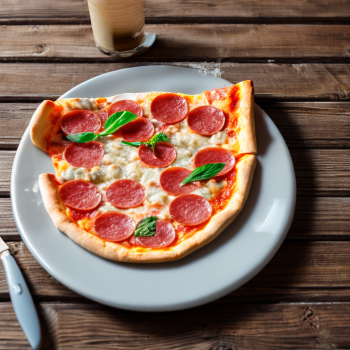}\hspace{-0.5mm}
    \includegraphics[width=\imgwidth]{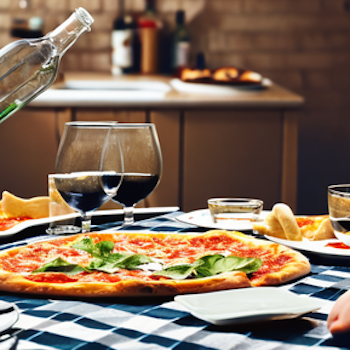}\\
    \vspace{1mm}
    \raisebox{0.5\imgwidth}{
        \begin{minipage}[c]{\txtwidth}
            \centering \footnotesize
            Snow next to road.
        \end{minipage}}\hspace{-0.5mm}
    \raisebox{0.5\imgwidth}{
        \begin{minipage}[c]{\txtwidth}
            \centering \footnotesize
            Snow next to road. A wintry landscape, where the glistening snow adds a touch of magic to the familiar thoroughfare.
        \end{minipage}}\hspace{-0.5mm}
    \raisebox{0.5\imgwidth}{
        \begin{minipage}[c]{\txtwidth}
            \centering \footnotesize
            Snow \textbf{on top of truck} next to road. \textbf{Sky, tree, sign, line, ground, building, car, cloud, and sidewalk on} road.
        \end{minipage}}\hspace{1.0mm}
    \includegraphics[width=\imgwidth]{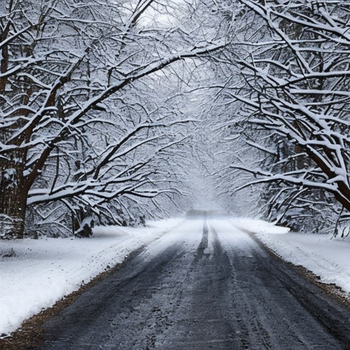}\hspace{-0.5mm}
    \includegraphics[width=\imgwidth]{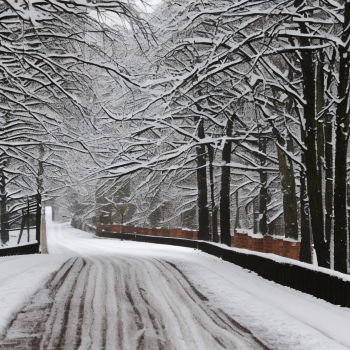}\hspace{-0.5mm}
    \includegraphics[width=\imgwidth]{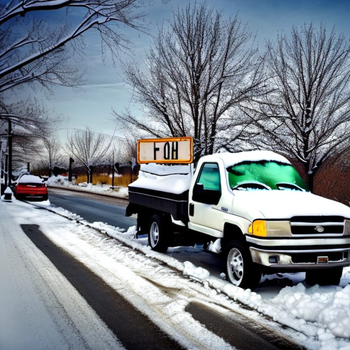}\\
    \vspace{1mm}
    \raisebox{0.5\imgwidth}{
        \begin{minipage}[c]{\txtwidth}
            \centering \footnotesize
            Car. Road near Street.
        \end{minipage}}\hspace{-0.5mm}
    \raisebox{0.5\imgwidth}{
        \begin{minipage}[c]{\txtwidth}
            \centering \footnotesize
            Car. Road near street. An everyday scene where the car navigates the road close to a charming, bustling street.
        \end{minipage}}\hspace{-0.5mm}
    \raisebox{0.5\imgwidth}{
        \begin{minipage}[c]{\txtwidth}
            \centering \footnotesize
            Car. \textbf{Sign on} road near street. \textbf{Bus and line on} street \textbf{near tree and building. Sky and sidewalk near building.}
        \end{minipage}}\hspace{1.0mm}
    \includegraphics[width=\imgwidth]{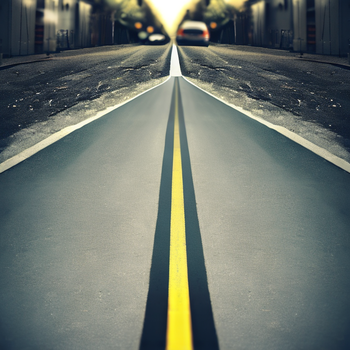}\hspace{-0.5mm}
    \includegraphics[width=\imgwidth]{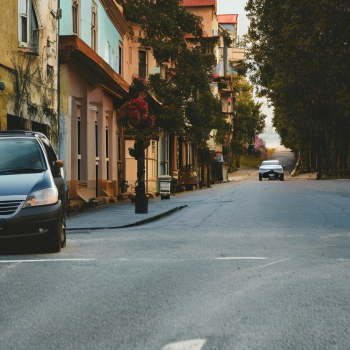}\hspace{-0.5mm}
    \includegraphics[width=\imgwidth]{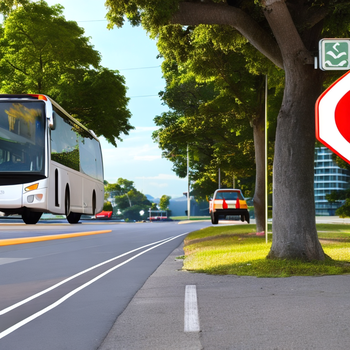}\\
    \vspace{\figmid}
    \caption{\emph{Qualitative comparison.} Visual Genome test examples showing input descriptions, enriched descriptions, and the resulting synthesized images. Column (a) shows the input description, columns (b) and (c) show the enriched descriptions from \emph{ChatGPT Direct} and our method, column (d) shows the image rendered from the input description, and columns (e) and (f) show the images rendered from the ChatGPT-enriched and our enriched descriptions, respectively. Bold text marks content added during enrichment.}
    \label{fig:samples}
    \vspace{\figdwn}
\end{figure*}
Fig.~\ref{fig:samples} compares the text-only \textit{ChatGPT Direct} baseline with our graph-based enrichment framework on selected VG test examples. Bold text marks content added during enrichment. Additional qualitative examples are provided in the Online Appendix (Fig.~\ref{fig:additional_samples}). The release package includes further qualitative examples, including strong enrichments and cases where graph-derived text can use generic predicates or weakly grounded object attachments.

For qualitative inspection, we randomly select subgraphs from the VG test split without constraining the number of objects or relations.
This subgraph strategy reflects the practical limitations of image generators when rendering scenes with many objects.
Thus, if the initial input description already contains a substantial number of objects, the image synthesizer cannot accurately depict the enriched scene with even more objects. This subgraph choice is an evaluation-time rendering constraint, not a constraint on the GCE formulation. Our proposed method carries out semantic enrichment iteratively, focusing on enriching one object and its relations at a time, regardless of the number of existing objects in the scene.
The image generator in \emph{Stage 2: Image Generation} is the fixed Stable Diffusion renderer used throughout the reported image experiments. Moreover, in \emph{Visual Scene Characterizer}, we employ a pre-trained residual CNN scene classifier \citep{zhou2017places}. During network training, gradients backpropagate through \emph{Stage 3: Training-Time Visual Alignment} and the fixed \emph{Stage 2: Image Generation} renderer; however, their parameters are not updated, because our objective function is designed for joint scene graph/image enrichment rather than for retraining those pre-trained modules for their original standalone tasks. The qualitative and quantitative visualizations in Fig.~\ref{fig:samples} and Table~\ref{tab:fid_is} are rendered with the same Stable Diffusion image generator from the simple or enriched textual descriptions.

\subsection{User Studies} \label{sec:user_studies}

\begin{table}
    \centering
    \vspace{\tabup}
    \caption{User studies with 110 participants. S1 compares Stable Diffusion and real images. S2 tests whether added content changes perceived realism. S3 tests whether the main input semantics are preserved. S4 compares simple and enriched outputs using the real image as reference.}
    \label{tab:user_study}
    \vspace{\tabmid}
    \resizebox{0.6\columnwidth}{!}{%
        \begin{tabular}{ccc}
            \hline
            \multicolumn{3}{c}{\textbf{S1:} Which one of the images is more realistic?}                                                                     \\
            \multicolumn{1}{c}{\textbf{(A) Stable Diffusion Image}} & \multicolumn{1}{c}{\textbf{(B) Real-World Image}} & \textbf{$\pm$ Standard Deviation} \\
            \multicolumn{1}{c}{20.09\%}                             & \multicolumn{1}{c}{79.91\%}                       & $\pm$ 23.27\%                     \\ \hline\hline
            \multicolumn{3}{c}{\textbf{S2:} Which one of the images is more realistic?}                                                                     \\
            \multicolumn{1}{c}{\textbf{(A) Enriched Image}}         & \multicolumn{1}{c}{\textbf{(B) Simple Image}}     & \textbf{$\pm$ Standard Deviation} \\
            \multicolumn{1}{c}{49.45\%}                             & \multicolumn{1}{c}{50.55\%}                       & $\pm$ 28.30\%                     \\ \hline\hline
            \multicolumn{3}{c}{\textbf{S3:} Does the image reflect the description?}                                                                        \\
            \multicolumn{1}{c}{\textbf{(A) Yes}}                    & \multicolumn{1}{c}{\textbf{(B) No}}               & \textbf{$\pm$ Standard Deviation} \\
            \multicolumn{1}{c}{70.82\%}                             & \multicolumn{1}{c}{29.18\%}                       & $\pm$ 13.92\%                     \\ \hline\hline
            \multicolumn{3}{c}{\textbf{S4:} With the real image as reference, which AI-generated image do you prefer?}                                      \\
            \multicolumn{1}{c}{\textbf{(A) Enriched Image}}         & \multicolumn{1}{c}{\textbf{(B) Simple Image}}     & \textbf{$\pm$ Standard Deviation} \\
            \multicolumn{1}{c}{83.40\%}                             & \multicolumn{1}{c}{16.60\%}                       & $\pm$ 12.24\%                     \\ \hline
        \end{tabular}%
    }
    \vspace{\tabdwn}
\end{table}

Human judgments complement the proxy graph and image-quality metrics by testing realism, description consistency, and preference for enriched versus direct generations. Because GCE is ill-posed, automatic scores against one GT target provide only partial evidence. Human judgment therefore complements the automatic proxy metrics by assessing whether the enriched outputs are realistic, description-consistent, and preferable to simple generated images. Table~\ref{tab:user_study} reports four human-judgment sections completed by 110 participants. All participants completed S1--S4 in order. S1, S2, and S3 each contain 10 items, while S4 contains 14 image triplets. The reported percentages aggregate responses across participants and items within each section, and the standard deviation summarizes variation over those participant-item responses for the corresponding section. The response records preserve the participant-item selections used to compute these summaries.

\textbf{S1: Real-Image vs. Stable-Diffusion Comparison.}
To assess how realistic images from the fixed image generator appear relative to real images, we presented users with ten pairs of images. Each pair consists of a real image selected from Visual Genome and an image generated by Stable Diffusion, both depicting similar scenes. Participants were asked: \emph{Which one is more realistic?} Participants selected the synthesized image as more realistic in only 20.09\% of comparisons. This result clarifies that our goal is not to make generated images indistinguishable from real photographs; the fixed renderer can still produce recognizable artifacts, and the central question is whether enrichment adds plausible scene content under the same generator.

\textbf{S2: Realism Preservation in Simple vs. Enriched Images.}
For ten simple descriptions extracted from subgraphs in the test split and their corresponding enriched descriptions produced by our model, pairs of images are synthesized. Users were asked: \emph{Which one is more realistic?} The enriched images contain more objects and richer object-relation content, which gives the fixed image generator more semantics to combine into one coherent image. Participants selected enriched and simple images at nearly equal rates. This supports the interpretation that graph-predicted additions preserve perceived realism under the same renderer: the added objects and relations remain compatible with the overall scene while enriching the generated content.

\textbf{S3: Description-to-Image Consistency.}
Ten pairs of simple descriptions, analogous to those in S2, were presented to participants alongside their corresponding enriched images. The question posed to the participants was: \emph{Does the generated image reflect the description?} Participants often judged the enriched images as reflecting the provided descriptions, which is consistent with the intended goal of adding content while preserving the main input semantics.

\textbf{S4: Real-Image-Guided Preference between Simple and Enriched Images.}
In this evaluation, 14 triplets of images were displayed to the users. Each triplet includes a pair of simple and enriched images, similar to S2, alongside their real-world VG images corresponding to identical scene descriptions. Users were explicitly informed which two images were AI-generated while specifying the real one. They were asked: \emph{Based on the real image, which AI-generated image do you prefer?} Participants more often selected the enriched images over the simple ones when both were compared against the corresponding real image; this pattern is consistent with the added content being perceived as meaningful while preserving scene coherence.

Taken together, these results support the feasibility of the framework under the reported Visual Genome setting. Because GCE admits multiple valid enrichments and several reported scores act as proxy measures, the quantitative results should be interpreted jointly with the qualitative examples and user studies.

\textbf{\emph{Reproducibility:}} The project release page is planned at \url{https://github.com/Mahdi-Naseri/GCE}. The release package includes the training, preprocessing, evaluation, graph-to-text rendering, repeated-seed aggregation, and configuration materials needed to reproduce the reported Visual Genome experiments.

\section{Conclusion}

In this paper, we addressed the task of \emph{Generated Contents Enrichment} by proposing an adversarial GCN-based framework with a multi-term objective. The framework enriches the scene representation underlying the input description and uses visual-domain signals during training to guide the generation of richer output images. More specifically, it seeks to narrow the gap between sparse descriptions and richer scene descriptions, while encouraging the added visual content to remain compatible with the original scene. The framework is designed to preserve the key scene elements specified in the input description. A pair of discriminators and alignment modules encourage the appended objects and relations to remain structurally and semantically coherent with the original scene. Overall, the results support GCE as a feasible direction for explicit scene enrichment before image generation, while broader graph-annotated datasets and stronger renderers remain natural extensions.

\bibliographystyle{unsrtnat}
\bibliography{refs}

\clearpage
\section*{Online Appendix}
\appendix

\begin{figure*}[t]
    \vspace{\figup}
    \centering
    \begin{minipage}{\imsize} \centering\scriptsize (a) Input Scene Graph \end{minipage}%
    \begin{minipage}{2\imsize} \centering\scriptsize (b) Enriched Scene Graph \end{minipage}%
    \begin{minipage}{\imsize} \centering\scriptsize (c) Input/\textbf{Enriched} Text \end{minipage}%
    \begin{minipage}{\imsize} \centering\scriptsize (d) Simple Image \end{minipage}%
    \begin{minipage}{\imsize} \centering\scriptsize (e) Enriched Image \end{minipage}\\
    \vspace{1mm}
    \raisebox{0.5\imsize}{
        \begin{minipage}[c]{\imsize}
            \centering \includegraphics[height=0.8\imsize]{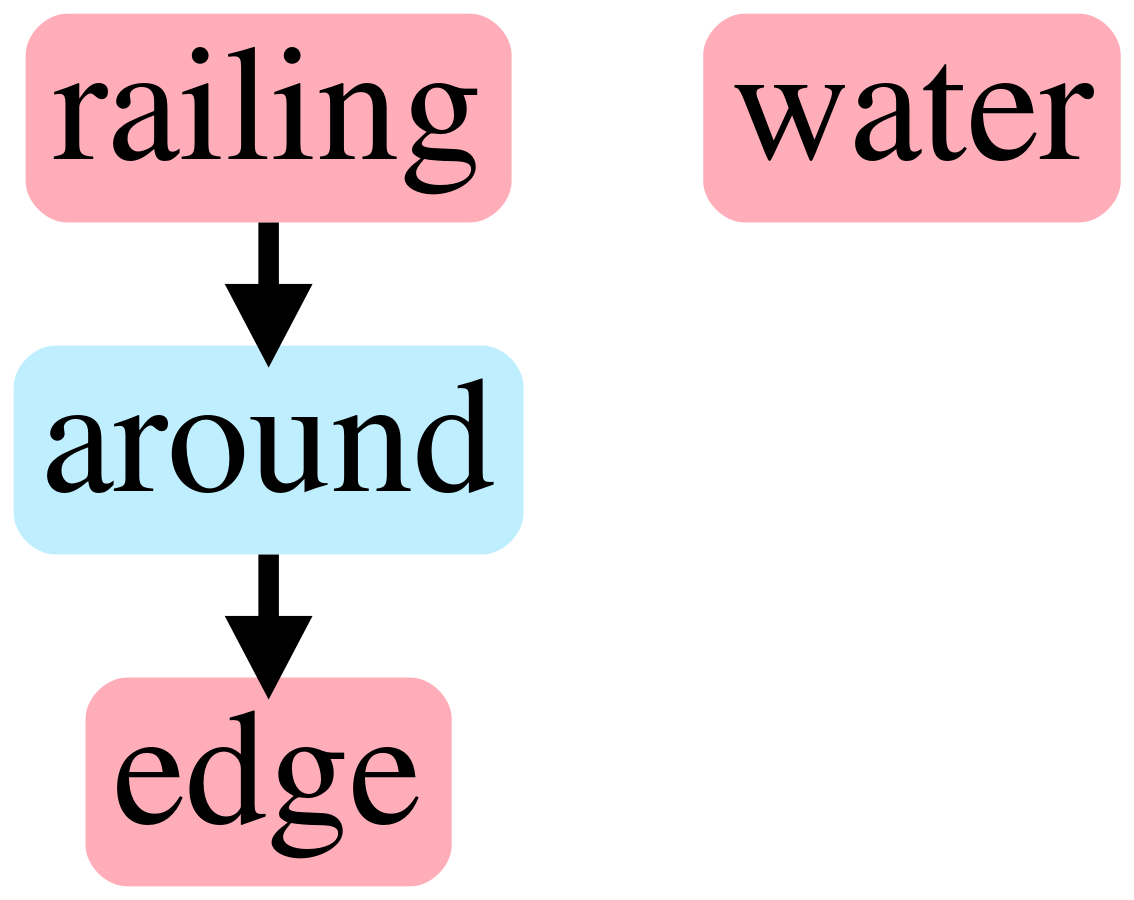}
        \end{minipage}}
    \raisebox{0.5\imsize}{
        \begin{minipage}[c]{2\imsize}
            \centering \includegraphics[height=\imsize]{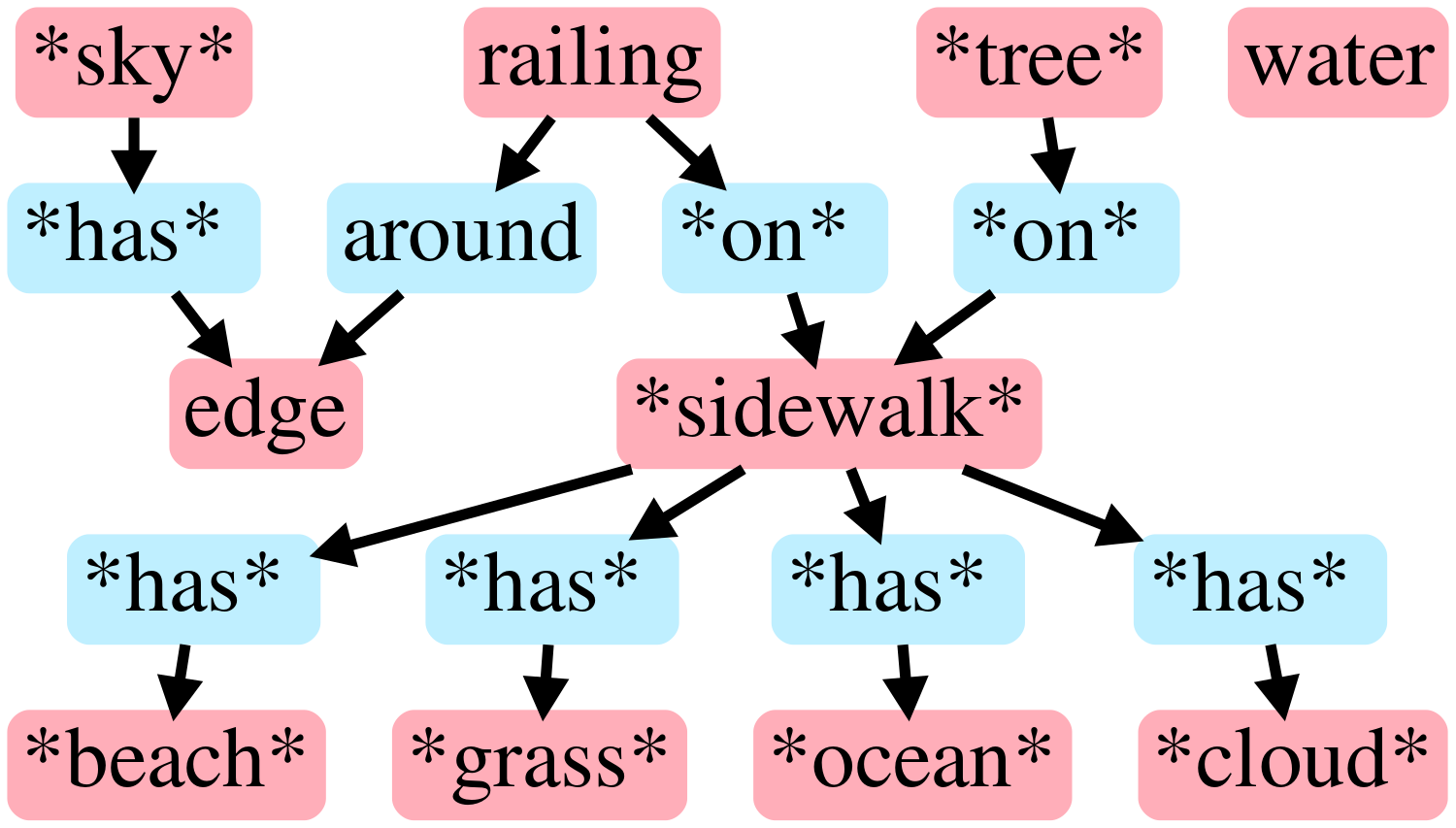}
        \end{minipage}}
    \raisebox{0.5\imsize}{
        \begin{minipage}[c]{\textsize}
            \centering \scriptsize Railing \textbf{on sidewalk} and around edge. Water. \textbf{ Sky has} edge. \textbf{Tree on sidewalk that has beach, ocean, grass, and cloud.}
        \end{minipage}}
    \includegraphics[width=\imsize]{figures/samples/railing/2.png}\hspace{-0.5mm}
    \includegraphics[width=\imsize]{figures/samples/railing/6.png}\\
    \vspace{1mm}
    \raisebox{0.5\imsize}{
        \begin{minipage}[c]{\imsize}
            \centering \includegraphics[height=0.8\imsize]{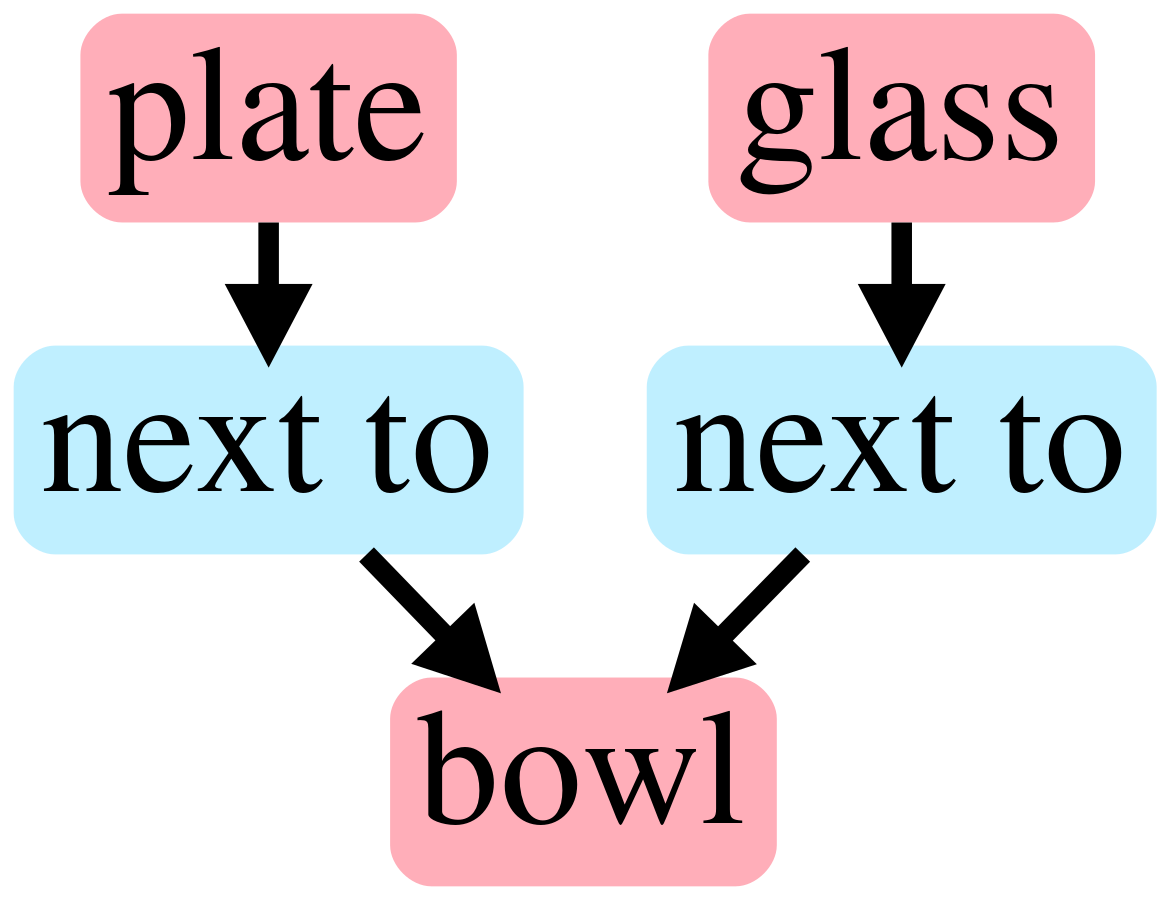}
        \end{minipage}}
    \raisebox{0.5\imsize}{
        \begin{minipage}[c]{2\imsize}
            \centering \includegraphics[width=\sgsize]{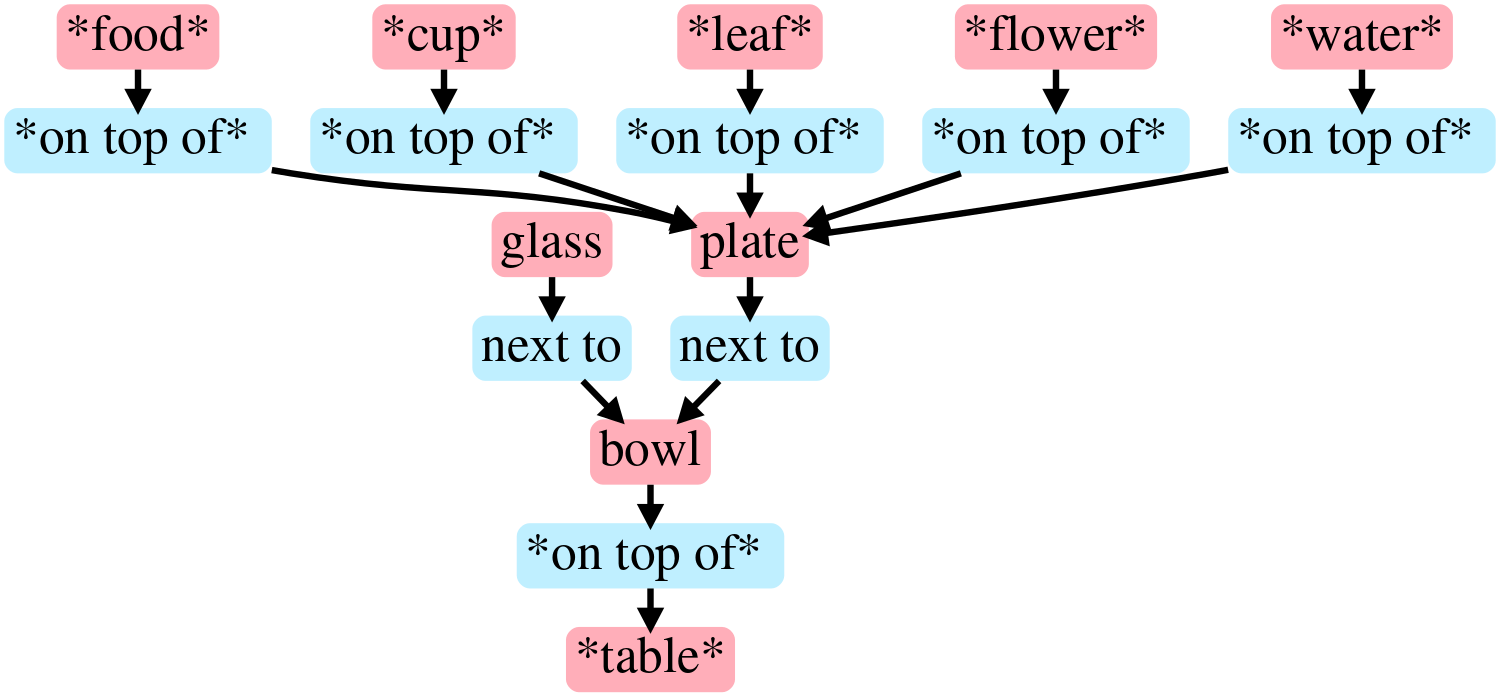}
        \end{minipage}}
    \raisebox{0.5\imsize}{
        \begin{minipage}[c]{\textsize}
            \centering \scriptsize Plate and glass next to bowl \textbf{on top of table. Food, cup, leaf, flower, water on top of} plate.
        \end{minipage}}
    \includegraphics[width=\imsize]{figures/samples/food_1/1.png}\hspace{-0.5mm}
    \includegraphics[width=\imsize]{figures/samples/food_1/5.png}\\
    \vspace{1mm}
    \raisebox{0.5\imsize}{
        \begin{minipage}{\imsize}
            \centering \includegraphics[width=\imsize]{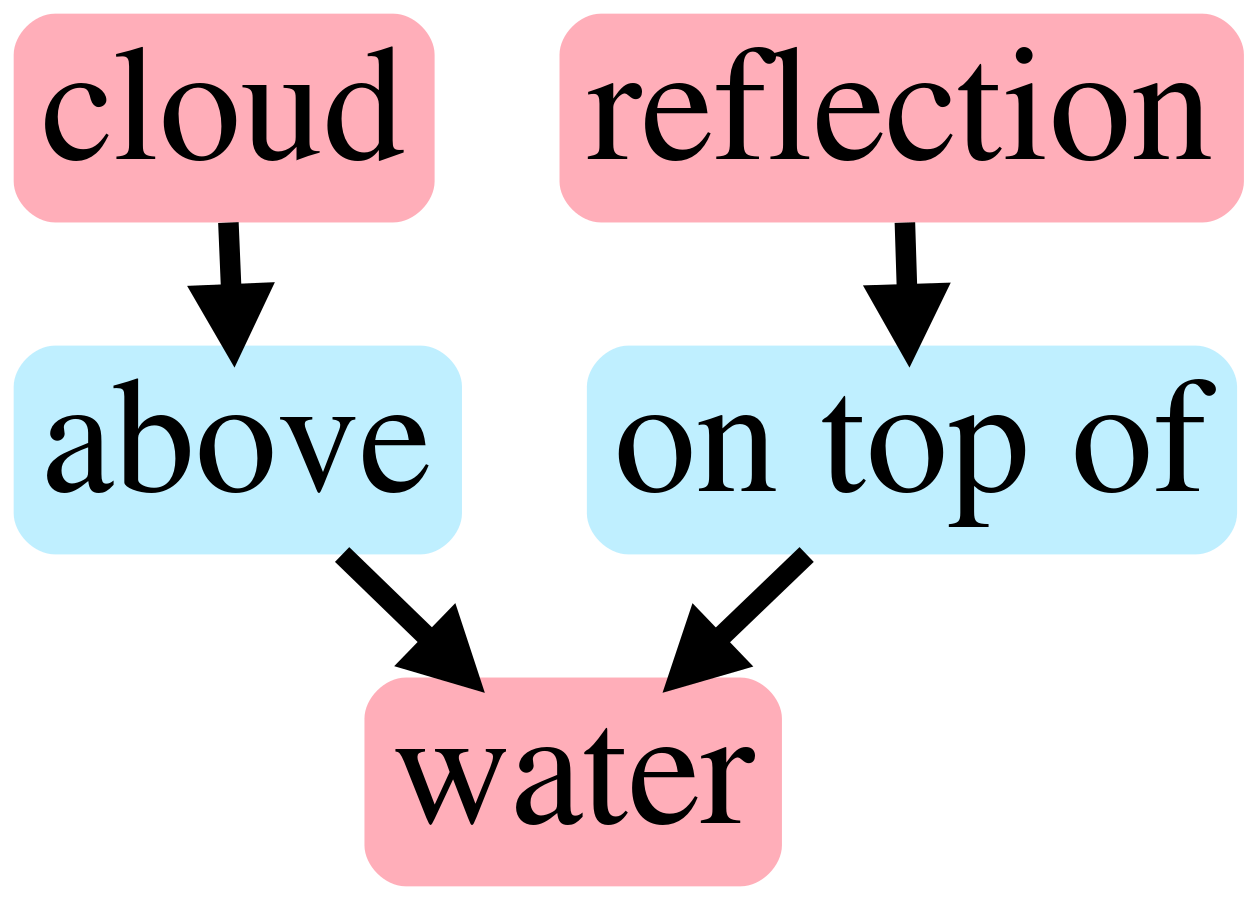}
        \end{minipage}}
    \raisebox{0.5\imsize}{
        \begin{minipage}{\sgsize}
            \centering \includegraphics[width=\sgsize]{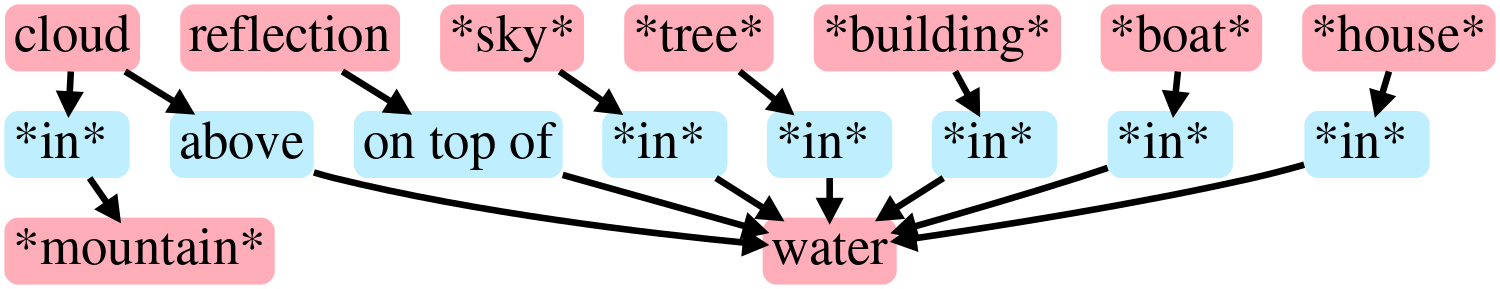}
        \end{minipage}}
    \raisebox{0.5\imsize}{
        \begin{minipage}[c]{\textsize}
            \centering \scriptsize Cloud \textbf{in mountain.} Cloud above water. Reflection on top of water. \textbf{Sky, tree, building, boat, and house in} water.
        \end{minipage}}
    \includegraphics[width=\imsize]{figures/samples/water/1.png}\hspace{-0.5mm}
    \includegraphics[width=\imsize]{figures/samples/water/6.png}\\
    \vspace{1mm}
    \raisebox{0.5\imsize}{
        \begin{minipage}{\imsize}
            \centering \includegraphics[width=\imsize]{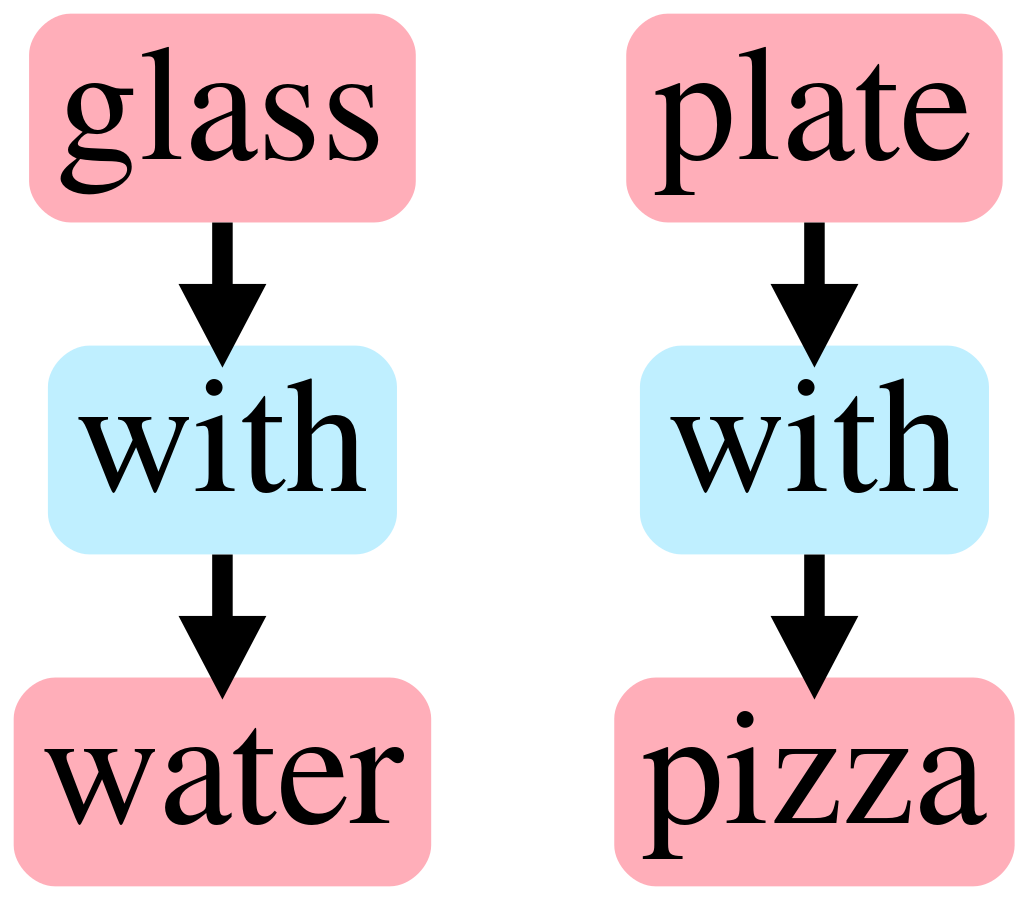}
        \end{minipage}}
    \raisebox{0.5\imsize}{
        \begin{minipage}{\sgsize}
            \centering \includegraphics[width=\sgsize]{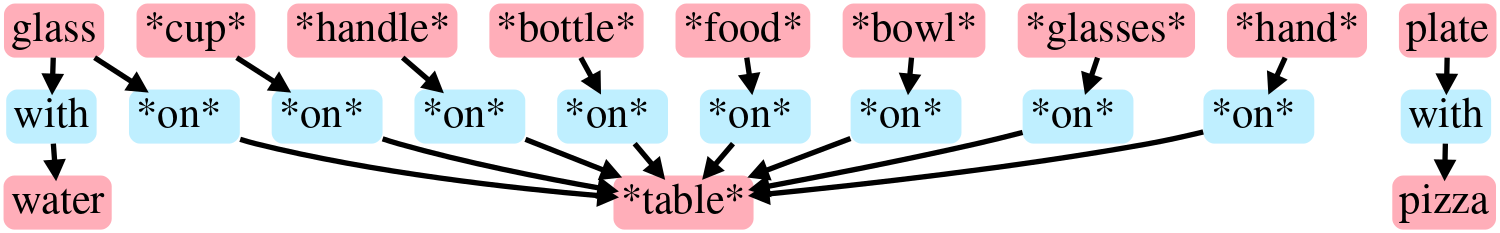}
        \end{minipage}}
    \raisebox{0.5\imsize}{
        \begin{minipage}[c]{\textsize}
            \centering \scriptsize Glass with water \textbf{on table.} Plate with pizza. \textbf{Cup, handle, bottle, food, bowl, glasses, and hand on table.}
        \end{minipage}}
    \includegraphics[width=\imsize]{figures/samples/food_2/1.png}\hspace{-0.5mm}
    \includegraphics[width=\imsize]{figures/samples/food_2/6.png}\\
    \vspace{1mm}
    \raisebox{0.5\imsize}{
        \begin{minipage}{\imsize}
            \centering \includegraphics[height=0.8\imsize]{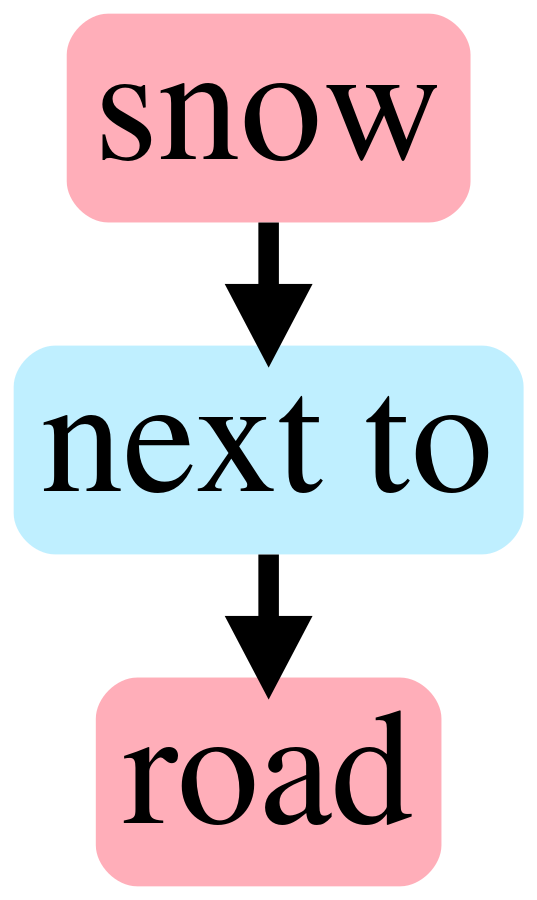}
        \end{minipage}}
    \raisebox{0.5\imsize}{
        \begin{minipage}{\sgsize}
            \centering \includegraphics[width=\sgsize]{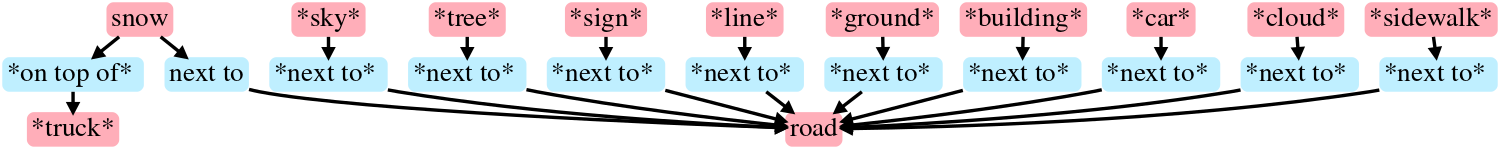}
        \end{minipage}}
    \raisebox{0.5\imsize}{
        \begin{minipage}[c]{\textsize}
            \centering \scriptsize Snow \textbf{on top of truck} next to road. \textbf{Sky, tree, sign, line, ground, building, car, cloud, and sidewalk on} road.
        \end{minipage}}
    \includegraphics[width=\imsize]{figures/samples/street_3/1.png}\hspace{-0.5mm}
    \includegraphics[width=\imsize]{figures/samples/street_3/7.png}\\
    \vspace{1mm}
    \raisebox{0.5\imsize}{
        \begin{minipage}{\imsize}
            \centering \includegraphics[width=\imsize]{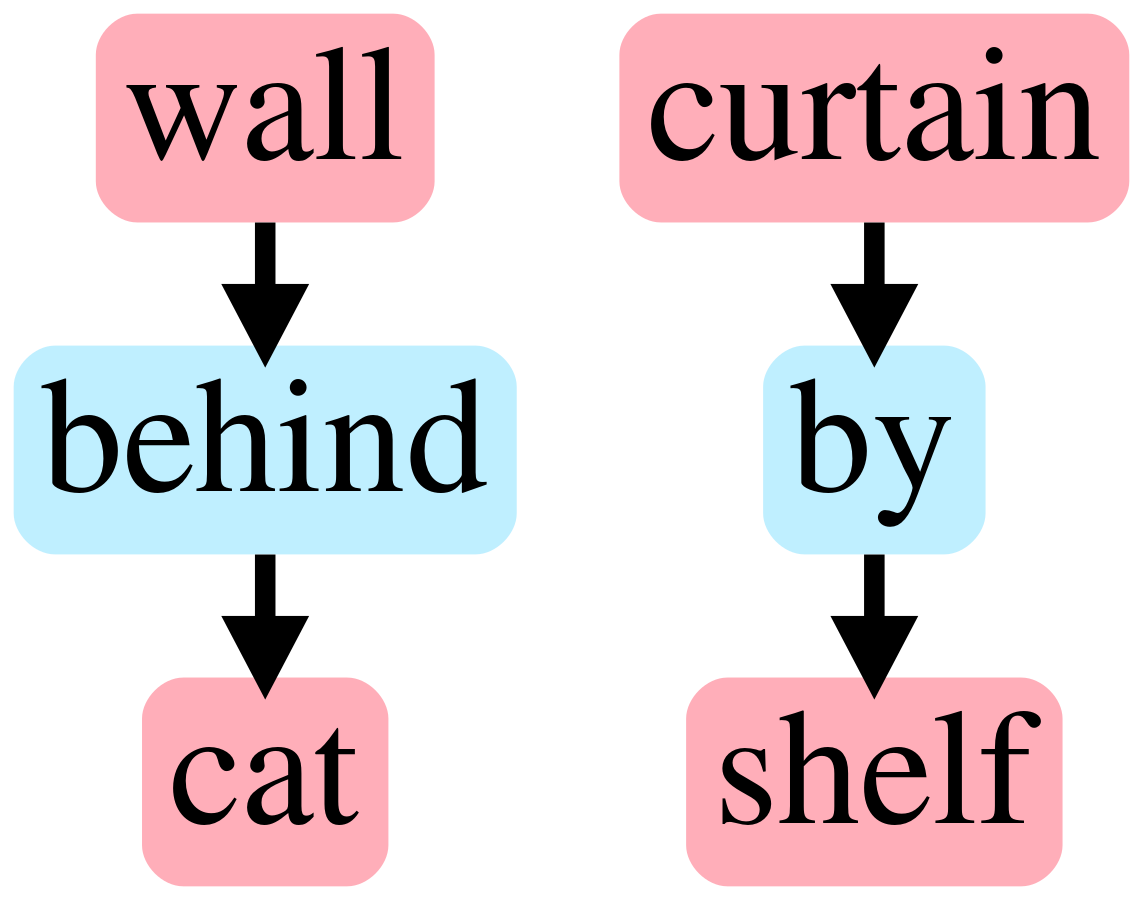}
        \end{minipage}}
    \raisebox{0.5\imsize}{
        \begin{minipage}{\sgsize}
            \centering \includegraphics[width=\sgsize]{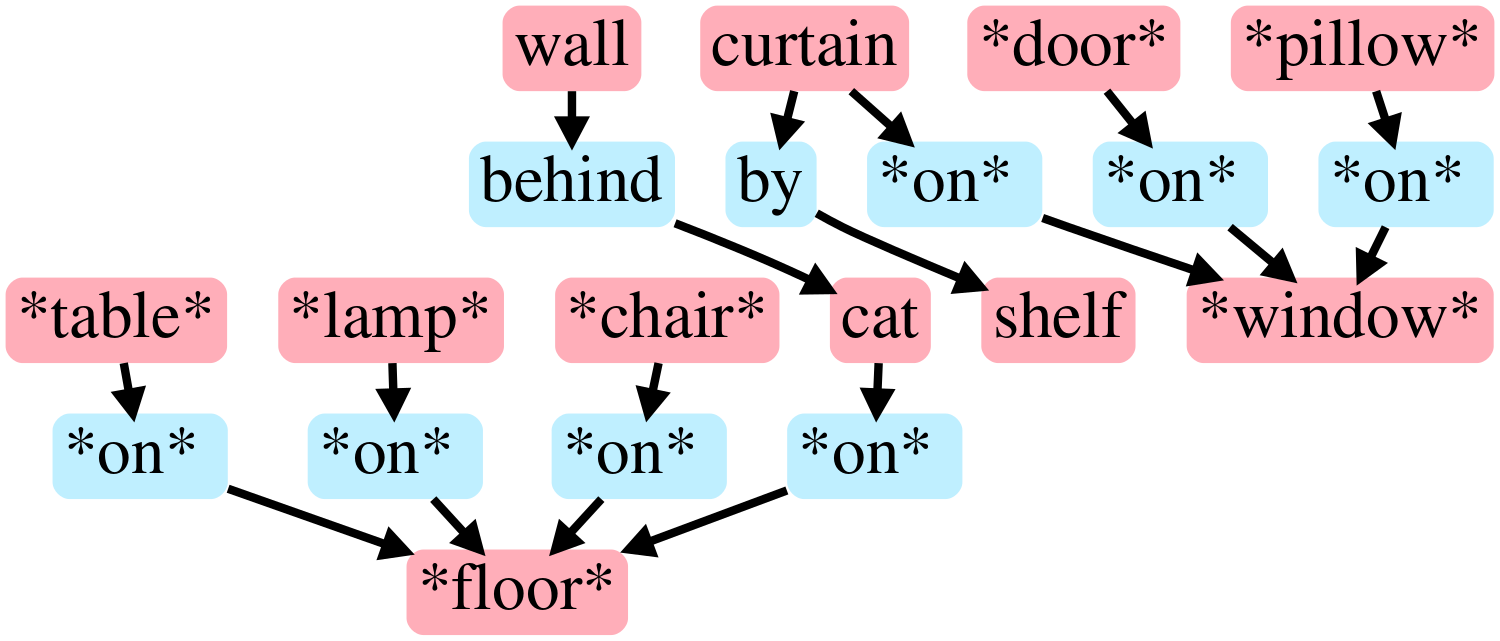}
        \end{minipage}}
    \raisebox{0.5\imsize}{
        \begin{minipage}[c]{\textsize}
            \centering \scriptsize Wall behind cat \textbf{on floor}. \textbf{Chair, lamp, and table on floor.} Curtain by shelf. Curtain \textbf{on window. Pillow on window. Door on window.}
        \end{minipage}}
    \includegraphics[width=\imsize]{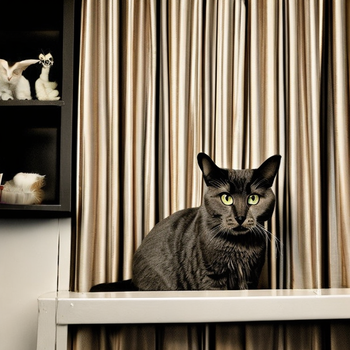}\hspace{-0.5mm}
    \includegraphics[width=\imsize]{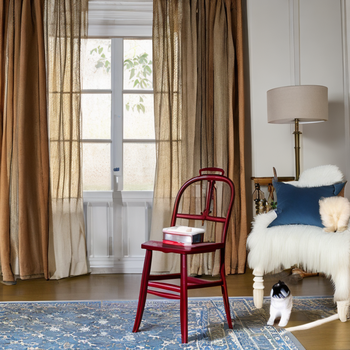}\\
    \vspace{\figmid}
    \caption{\emph{Additional qualitative results.} Additional Visual Genome examples in the Online Appendix. Each row follows the same five-part format: input graph, enriched graph, enriched text, simple image, and enriched image. Bold text marks content added during enrichment.}
    \label{fig:additional_samples}
    \vspace{\figdwn}
\end{figure*}

\section{Supplementary Details} \label{sec:app_method}

\subsection{GConv Operation} \label{sec:app_gconv}

Each GConv layer updates relation-triplet representations before node-level aggregation. For a triplet \((o_i,r,o_j)\), the subject, predicate, and object vectors \((v_i,v_r,v_j)\) are concatenated, mapped through fully connected layers, and divided into candidate subject, edge, and object updates \((\bar v_i,\bar v_r,\bar v_j)\). Candidate node updates are averaged over all incident triplets, passed through additional fully connected layers, and combined with a skip connection before the next graph-convolution layer.
\subsection{GCN Implementation} \label{sec:app_gcn_impl}

For mini-batches, relation triplets from all graphs are processed in parallel. Following \citep{johnson2018image}, a dummy image node and image-link predicates keep scene graphs separated inside a mini-batch. The \emph{Enriching Edge Detector} receives the object nodes and hidden object vectors, maps each node to subject-side and object-side embeddings with two separate MLPs, and scores directed candidate edges by inner products between those embeddings.

\subsection{Generator GCN Architectures} \label{sec:app_generator_gcn_arch}
The generator uses two GCNs with the same architecture and separate weights. The selected architecture increases the embedding dimension to process graph context, reduces it to encourage compact encoded information, and returns to the output dimension. The final configuration uses object and predicate embedding layers, two GCN classifier layers, and two enriching-edge layers.

\subsection{CLIP Text Encoder} \label{sec:app_clip_text}

The image-text aligner uses CLIP ViT-B/32. Because CLIP encodes text, each scene graph is first linearized into graph-derived text. For the CLIP auxiliary path, the implementation filters unavailable predicates, formats relation phrases as subject--predicate--object strings, includes object labels that do not participate in a relation, de-duplicates the phrase lists, shuffles standalone-object and relation phrases under the process random state, and truncates the resulting text to the 75-word CLIP budget. The same graph-to-text conversion procedure is used for the CLIP alignment path and for the graph-derived text prompts rendered by Stable Diffusion; renderer-specific settings are reported in Appendix Section~\ref{sec:app_image_generator}.

For the training pathway denoted by \(\phi(g)\) in Eq.~\ref{eq:img_synth}, graph predictions are represented before category selection as continuous conditioning embeddings. For an object distribution \(p\) over the object-label vocabulary with embedding matrix \(E_{\mathrm{obj}}\), the conditioning vector is \(e_{\mathrm{obj}}=p^{\top}E_{\mathrm{obj}}\); predicate distributions are handled analogously with \(E_{\mathrm{pred}}\). This continuous relaxation replaces selecting a single embedding row and permits gradients from the visual and image-text objectives to flow to the object and predicate logits.

\subsection{ChatGPT Prompt Templates} \label{sec:app_chatgpt_prompts}
For the text-based baseline variants, the input description was inserted into one of the following prompt templates. These prompts define short out-of-the-box prompt-expansion baselines. They are not optimized LLM prompt-engineering systems.
\begin{enumerate}
    \item \textit{ChatGPT Direct:} Please complete this given text so that the final text contains between 15 and 25 words: ``\textit{input\_description}''.
    \item \textit{ChatGPT Scene:} Please complete this given text \textbf{describing a scene} so that the final text contains between 15 and 25 words: ``\textit{input\_description}''.
    \item \textit{ChatGPT Object:} Please complete this given text \textbf{describing a scene with more related objects} so that the final text contains between 15 and 25 words: ``\textit{input\_description}''.
\end{enumerate}

\subsection{Image Generator} \label{sec:app_image_generator}
Our image generation module uses a fixed Stable Diffusion renderer \citep{rombach2022high}. The rendering implementation uses the Diffusers \texttt{StableDiffusionPipeline} with checkpoint \texttt{stabilityai/stable-diffusion-2-1}, fp16 CUDA inference, and guidance-scale candidates 7.5, 9, 15, and 25. The rendering scripts do not explicitly override the scheduler, denoising step count, negative prompt, or diffusion generator seed, so those settings follow pipeline defaults and saved image artifacts rather than a separately controlled seed protocol. Each simple, GT, ChatGPT-enriched, or GCE-enriched scene representation is converted to text before rendering, and the same Stable Diffusion checkpoint family is used for the reported image-quality comparisons and user-study images. The renderer and encoder parameters are held fixed during GCE training, so gradients update the scene graph enrichment components rather than fine-tuning the text-to-image, scene-classifier, or CLIP models.
\subsection{Extended Proxy Scene Graph Enrichment Metrics}
\label{sec:app_expanded_proxy_metrics}

The main text introduces the motivation for proxy scene graph enrichment metrics: GCE can have several plausible enrichments for the same sparse input, so exact recovery of one annotated target is informative but not exhaustive. This appendix gives formal definitions for the metrics used in Table~\ref{tab:expanded_setting_a_canonical_full}.

\paragraph{Metric definitions.}
We define the metrics in the same order as the main table: object accuracy, available-edge accuracy, no-edge accuracy, available-predicate accuracy, triplet precision/recall/F1, and MEP+.
Let \(\mathcal{C}\) be the object label set and \(\mathcal{R}\) the predicate label set after canonical normalization. Let \(K\) be the set of single-object removal test cases. Each case \(k\in K\) is one reduced seed graph \(g_k=(O_k,E_k)\) created by removing a target object \(o_k^\star\in\mathcal{C}\) from a full Visual Genome graph. Let \(T_k\subseteq\mathcal{C}\times\mathcal{R}\times\mathcal{C}\) be the set of scored ground truth triplets incident to \(o_k^\star\). The method predicts an added object \(\hat o_k\) and a set of incident triplets \(\hat T_k\). All per-case fractions below are defined as zero when their denominator is zero.

Object accuracy is computed over all test cases:
\begin{equation}
    A_{\mathrm{obj}}=\frac{1}{|K|}\sum_{k\in K}\mathbf{1}[\hat o_k=o_k^\star].
    \label{eq:metric_object_accuracy}
\end{equation}
Let \(V=\{k\in K: |T_k|>0\}\) be the visible-target cases, meaning cases where the removed object has at least one scored ground truth incident triplet. A predicted triplet is correct only when its subject label, predicate label, and object label match a ground truth triplet. The scorer also gates triplet true positives by object recovery:
\begin{equation}
    c_k=\mathbf{1}[\hat o_k=o_k^\star]\,|\hat T_k\cap T_k|.
    \label{eq:metric_triplet_count}
\end{equation}
The object gate is intentional: relation credit is awarded only for triplets attached to the correctly recovered added object. Although the triplet intersection also checks object labels inside each triplet, the explicit gate prevents a method from receiving relation credit after choosing the wrong added object identity.
For \(k\in V\), define \(P_k=c_k/|\hat T_k|\), \(R_k=c_k/|T_k|\), and \(F_k=2P_kR_k/(P_k+R_k)\) with the zero-denominator convention above. The reported triplet metrics are macro averages over visible-target cases:
\begin{align}
    P_{\mathrm{tri}} & = \frac{1}{|V|}\sum_{k\in V}P_k,\label{eq:metric_triplet_precision} \\
    R_{\mathrm{tri}} & = \frac{1}{|V|}\sum_{k\in V}R_k,\label{eq:metric_triplet_recall}    \\
    F_{\mathrm{tri}} & = \frac{1}{|V|}\sum_{k\in V}F_k.\label{eq:metric_triplet_f1}
\end{align}
Thus \(F_{\mathrm{tri}}\) is the mean of per-case F1 values, not the harmonic mean of the already averaged precision and recall. Tables display \(100P_{\mathrm{tri}}\), \(100R_{\mathrm{tri}}\), and \(100F_{\mathrm{tri}}\), which is why values such as \(2.08\) correspond to \(0.0208\) on the underlying \([0,1]\) scale.

For edge and predicate metrics, the evaluator uses label-level slots rather than instance-ID-level slots. Let \(\bar O_k\) be the set of normalized object labels appearing in the seed graph. For each label \(u\in\bar O_k\), there are two directional edge slots involving the held-out object: \(u\rightarrow o_k^\star\) and \(o_k^\star\rightarrow u\). A slot is active when a triplet with that direction is present. Let \(\Omega_k\) be this set of slots, \(S_k\subseteq\Omega_k\) the ground truth active slots where an incident edge exists, and \(\hat S_k\subseteq\Omega_k\) the predicted active slots. A single predicted triplet has one predicate label. We use set notation only because, after label normalization, multiple triplets with different instance IDs can collapse into the same directional label-level slot. Therefore, \(\Pi_k(s)\) and \(\hat\Pi_k(s)\) denote the ground truth and predicted predicate-label sets observed for slot \(s\). Available-edge accuracy, available-predicate accuracy, and no-edge accuracy are
\begin{equation}
    A_{\mathrm{edge}}=\frac{1}{|V|}\sum_{k\in V}\frac{|S_k\cap\hat S_k|}{|S_k|},
    \label{eq:metric_available_edge}
\end{equation}
\begin{equation}
    A_{\mathrm{pred}}=\frac{1}{|V|}\sum_{k\in V}
    \frac{\sum_{s\in S_k\cap\hat S_k}\mathbf{1}[\Pi_k(s)\cap\hat\Pi_k(s)\neq\emptyset]}{|S_k\cap\hat S_k|},
    \label{eq:metric_available_predicate}
\end{equation}
\begin{equation}
    A_{\mathrm{noedge}}=\frac{1}{|V|}\sum_{k\in V}
    \frac{|\{s\in\Omega_k: s\notin S_k \ \mathrm{and}\ s\notin \hat S_k\}|}{|\{s\in\Omega_k: s\notin S_k\}|}.
    \label{eq:metric_noedge_accuracy}
\end{equation}
In words, \(A_{\mathrm{edge}}\) measures how many ground truth incident edge slots are activated, \(A_{\mathrm{pred}}\) measures predicate correctness only over slots that are active in both \(S_k\) and \(\hat S_k\), and \(A_{\mathrm{noedge}}\) measures the true negative rate over directional slots without a ground truth incident edge. The \(A_{\mathrm{pred}}\) denominator uses \(|S_k\cap\hat S_k|\) rather than \(|S_k|\) so that missed active slots are counted by edge recall rather than counted again as predicate errors.

MEP+ adapts the modified edge precision idea from GEMS \citep{agarwal2023gems} from whole-graph expansion to one-step GCE triplets. The underlying score lies in \([0,1]\), with higher values indicating that more predicted added triplets are supported by similar training cases. For each test case, let \(B_k\) be the set of training triplets retrieved as plausible support for the reduced seed graph. The scorer first retrieves triplets from training cases with the exact same reduced graph signature, meaning the same normalized seed object labels and seed triplets. If this set is empty, it falls back to training triplets with the same normalized seed object-label set. Thus, \(B_k\) is searched separately for each test case and acts as evidence for what additions were observed in comparable training contexts. The per-case and average scores are
\begin{equation}
    M_k^+=\frac{|\hat T_k\cap B_k|}{|\hat T_k|},\qquad
    A_{\mathrm{MEP}^{+}}=\frac{1}{|V|}\sum_{k\in V}M_k^+.
    \label{eq:metric_mep_plus}
\end{equation}
MEP+ complements the recovery metrics: triplet precision, recall, and F1 test agreement with the held-out annotated structure, while MEP+ tests whether the predicted added triplets have support in comparable Visual Genome training contexts.

\subsection{Pair of Discriminators} \label{sec:app_pair_discriminators}

On Visual Genome with the selected architectures, distinguishing enriched scene graphs from original data is usually easier than the enrichment task itself. Consequently, the discriminators can become much stronger than the generator unless their capacity or update frequency is controlled. We tune discriminator capacity with embedded dimensions from 16 to 256 and tune the discriminator update frequency; the selected final configuration uses a smaller discriminator embedding dimension than the generator and updates the discriminators less frequently than every generator step.

\subsection{Training Details \& Hyperparameter Tuning} \label{sec:app_training_details}

The final selected configuration uses mini-batches of 32 graphs, GConv dropout 0.1, no GConv normalization, LeakyReLU activations, generator embedding dimension 256, GCN architecture \((1,1,1,1,1)\), generator FC dropout 0.1 with batch normalization and LeakyReLU, two classifier layers, two enriching-edge layers, discriminator embedding dimension 16, discriminator architecture \((1,\frac{1}{2},\frac{1}{4},\frac{1}{8},\frac{1}{8},\frac{1}{8})\), and discriminator updates every 200 generator steps. The selected loss weights are 1000 for object classification, 100 for available-predicate classification, 0.1 for no-relation predicate classification, 1 for edge prediction, 0.1 for the GAN loss, 200 for the scene-feature loss, and the \(hpooled_{l_1}\) scene-feature option.

To evaluate iterative enrichment, the model is required to add an object not already present in the input scene graph at each inference step. We use Adam \citep{kingma2014adam} with a learning rate of \(10^{-4}\). We use early stopping based on validation-split accuracy. On the Visual Genome dataset, continued training improves training metrics while worsening validation metrics, indicating overfitting.

Before joint training with visual and image-text alignment losses, we first optimize only the scene graph generator and discriminators, without the image synthesizer, because including image synthesis substantially increases training and inference cost. Loss weights also have a strong effect on performance. For example, assigning a higher weight to \emph{No-Relation Pred.} can produce a model that predicts an enriching object without useful enriching edges.

Candidate models were ranked using validation-split proxy metrics, and qualitative inspection on a limited set of scene graphs was used as a secondary sanity check before selecting the final configuration. Because the loss weights and architecture choices create a large search space, we select a limited set of configurations using Bayesian hyperparameter optimization and train multiple candidate instances. The final reported configuration uses the \(L_1\) feature-distance form of \(\LL_{sc}\) in the main method.

\subsection{Reproducibility Package}
For reproducibility, the release package at \url{https://github.com/Mahdi-Naseri/GCE} includes the Visual Genome preprocessing interface, reduced-graph evaluator, selected hyperparameter configuration, metric scripts, prompt and image-generation artifacts, repeated-seed aggregation scripts, and commands needed to reproduce the reported proxy-metric tables from saved predictions or regenerated outputs.
The release package preserves the full tuning records and additional qualitative examples that were omitted from this length-limited appendix.

\end{document}